%% file: main.tex
\documentclass[nohyperref]{article}

\usepackage{microtype}
\usepackage{graphicx}
\usepackage{subfigure}
\usepackage{booktabs} 

\usepackage{graphicx}
\usepackage{multirow, multicol}
\usepackage{adjustbox}
\usepackage{arydshln}

\usepackage{caption}

\usepackage{algorithm,algorithmic}
\usepackage{extarrows}
\usepackage{bbm}
\usepackage{hyperref} 

\usepackage{hyperref}

\usepackage[accepted]{icml2023}

\usepackage{amsmath}
\usepackage{amssymb}
\usepackage{mathtools}
\usepackage{amsthm}

\usepackage[capitalize,noabbrev]{cleveref}

\theoremstyle{plain}
\newtheorem{theorem}{Theorem}[section]

\newtheorem{lemma}[theorem]{Lemma}

\theoremstyle{definition}

\newtheorem{assumption}[theorem]{Assumption}
\theoremstyle{remark}
\newtheorem{remark}[theorem]{Remark}

\input{ANNOTATIONS.tex}

\input{COMMANDS.tex}

\ExplSyntaxOn
\cs_set:Npn \loadimage #1#2#3#4 {
    \seq_clear:N \l_tmpa_seq
    \int_step_inline:nnn {#3} {#3 + #4 - 1} {
        \seq_put_right:Nx \l_tmpa_seq {
            \exp_not:N \includegraphics
                [\exp_not:n {#2}]
                {#1/##1.png}
        }
    }
    \seq_use:Nn \l_tmpa_seq {&}
}

\cs_set:Npn \calctotalwidth #1#2 {
    \fp_eval:n {
        (#2) * (#1) + (#2 - 1) * (\tabcolsep)
    } pt
}
\ExplSyntaxOff

\icmltitlerunning{Optimization for Amortized Inverse Problems}

\begin{document}

\twocolumn[
\icmltitle{
Optimization for Amortized Inverse Problems
}



\icmlsetsymbol{equal}{*}

\begin{icmlauthorlist}
\icmlauthor{Tianci Liu}{purdue}
\icmlauthor{Tong Yang}{pku}
\icmlauthor{Quan Zhang}{msu}
\icmlauthor{Qi Lei}{nyu}
\end{icmlauthorlist}

\icmlaffiliation{purdue}{School of Electrical and Computer Engineering, Purdue University, United States. email: \href{mailto:liu3351@purdue.edu}{\texttt{liu3351@purdue.edu}}}
\icmlaffiliation{pku}{Center of Data Science, Peking University, China. email: \href{mailto:tongyang@stu.pku.edu.cn}{\texttt{tongyang@stu.pku.edu.cn}}}
\icmlaffiliation{msu}{Department of Accounting and Information Systems, Michigan State University, United States. email: \href{mailto:quan.zhang@broad.msu.edu}{\texttt{quan.zhang@broad.msu.edu}}}
\icmlaffiliation{nyu}{Courant Institute of Mathematical Sciences and Center for Data Science,
New York University, United States. email: \href{mailto:ql518@nyu.edu}{\texttt{ql518@nyu.edu}} }



\vskip 0.3in
]



\printAffiliationsAndNotice{}  

\begin{abstract}
\input{subfiles/0_abstract}
\end{abstract}

\input{subfiles/1_intro}

\input{subfiles/2_related_work}
\input{subfiles/3_method}

\input{subfiles/4_theorem}
\input{subfiles/5_experiment}

\input{subfiles/6_conclusion}

\bibliography{ref}
\bibliographystyle{icml2023}

\newpage
\appendix
\onecolumn
\input{subfiles/7_appendix}



\end{document}

%% file: ANNOTATIONS.tex
\usepackage{ifthen}
\usepackage[normalem]{ulem} 
\usepackage{xcolor}
\usepackage{amssymb}

\newboolean{showedits}
\setboolean{showedits}{true} 
\ifthenelse{\boolean{showedits}}
{
\newcommand{\del}[1]{\textcolor{red}{\sout{#1}}} 
}{
\newcommand{\del}[1]{} 

}

\newboolean{showcomments}
\setboolean{showcomments}{true} 
\newcommand{\id}[1]{$-$Id: scgPaper.tex 32478 2010-04-29 09:11:32Z oscar $-$}

\ifthenelse{\boolean{showcomments}}
{\newcommand{\nbc}[3]{
{\colorbox{#3}{\bfseries\sffamily\scriptsize\textcolor{white}{#1}}}
{\textcolor{#3}{\sf\small$\blacktriangleright$\textit{#2}$\blacktriangleleft$}}}
}
{\newcommand{\nbc}[3]{}
\renewcommand{\del}[1]{} 
}

\definecolor{ibcolor}{rgb}{0.9,0.5,0}
\definecolor{dsrcolor}{rgb}{0.5,0.6,0}
\definecolor{cfcolor}{rgb}{0,0.5,0.9}
\definecolor{lwcolor}{rgb}{0.2,0.8,0.4}
\definecolor{eycolor}{rgb}{0.7,0.6,1.0}
\definecolor{oldcolor}{rgb}{0.2,0.2,0.2}
\definecolor{tdcolor}{rgb}{0.0,0.5,0.7}

%% file: COMMANDS.tex
\newcommand{\beq}{\vspace{0mm}\begin{equation}}
\newcommand{\eeq}{\vspace{0mm}\end{equation}}
\newcommand{\beqs}{\vspace{0mm}\begin{eqnarray}}
\newcommand{\eeqs}{\vspace{0mm}\end{eqnarray}}
\newcommand{\barr}{\begin{array}}
\newcommand{\earr}{\end{array}}

\newcommand{\Amat}{{\bf A}}

\newcommand{\Imat}{{\bf I}}

\newcommand{\ev}{{\boldsymbol{e}}}

\newcommand{\xv}{{\boldsymbol{x}}}
\newcommand{\yv}{{\boldsymbol{y}}}
\newcommand{\zv}{{\boldsymbol{z}}}

\newcommand{\given}{\,|\,}

\newcommand{\lambdav}{{\boldsymbol{\lambda}}}

\newcommand{\xiv}{{\boldsymbol{\xi}}}

\newcommand{\zeros}{\boldsymbol{0}}

\newcommand{\R}{{\mathbb{R}}}
\newcommand{\E}{{\mathbb{E}}}
\newcommand{\N}{{\mathbb{N}}}

\usepackage{booktabs}
\usepackage{array}
\newcolumntype{L}[1]{>{\raggedright\let\newline\\\arraybackslash\hspace{0pt}}m{#1}}
\newcolumntype{C}[1]{>{\centering\let\newline\\\arraybackslash\hspace{0pt}}m{#1}}
\newcolumntype{R}[1]{>{\raggedleft\let\newline\\\arraybackslash\hspace{0pt}}m{#1}}

\newcommand{\loss}{{\cal L}}

\newcommand{\gaus}{{\cal N}}
\newcommand{\norm}[1]{\left\| #1 \right\|}

\DeclareMathOperator*{\argmin}{argmin}
\DeclareMathOperator*{\argmax}{argmax}

\DeclareMathOperator*{\proj}{Proj}
\DeclareMathOperator*{\clip}{clip}

\newcommand{\notion}{AIPO}

%% file: subfiles/0_abstract.tex
Incorporating a deep generative model as the prior distribution in inverse problems has established substantial success in  reconstructing images from corrupted observations. Notwithstanding, the existing optimization approaches use gradient descent largely without adapting to the non-convex nature of the problem and can be sensitive to initial values, impeding further performance improvement. In this paper, we propose an efficient amortized optimization scheme for inverse problems with a deep generative prior.
Specifically, the optimization task with high degrees of difficulty is decomposed into optimizing a sequence of much easier ones. We provide a theoretical guarantee of the proposed algorithm and empirically validate it on different inverse problems. As a result, our approach outperforms baseline methods qualitatively and quantitatively by a large margin.
  

%% file: subfiles/1_intro.tex
\section{Introduction}
\looseness -1 Inverse problems
aim to reconstruct the true image/signal $\xv_{T}$ from a corrupted (noisy or lossy) observation  $$ \yv = f(\xv_T) + \ev,$$
where $f$ is a known forward operator and $\ev$ is the noise. 
The problem is reduced to denoising  when $f(\xv) = \xv$ is the identity map and is reduced to a compressed sensing~\citep{candes2006stable,donoho2006compressed}, inpainting~\citep{vitoria2018semantic}, or a super-resolution problem~\citep{menon2020pulse} when $f(\xv) =\Amat \xv$ and $\Amat$ maps $\xv$ to an equal or lower dimensional space.

Inverse problems are generally ill-posed in the sense that there may exist infinitely many possible solutions, and thus require some natural signal priors to reconstruct the corrupted image~\citep{ongie2020deep}. 
Classical methods assume smoothness, sparsity in some basis, or other geometric properties on the image structures~\citep{candes2006stable, donoho2006compressed, danielyan2011bm3d, yu2011solving}. 
However, 
such assumptions may be too general and not task-specific. 
Recently, deep generative models, such as the 
generative adversarial network (GAN) and its variants~\citep{goodfellow2014generative, karras2017progressive, karras2019style}, are used as the prior of inverse problems after pre-training and have established great success \citep{bora2017compressed,hand2018global, hand2018phase,asim2020blind,jalal2020robust, jalal2021robust}.
Compared to classical methods, using a GAN prior is able to produce better reconstructions at much fewer measurements~\citep{bora2017compressed}.

\citet{asim2020invertible} points out that a GAN prior can be prone to representation errors and significant performance degradation if the image to be recovered is out of the data distribution where the GAN is trained. 
To address this limitation, the authors propose replacing the GAN prior with normalizing flows (NFs)~\cite{rezende2015variational}. 
NFs are invertible generative models that learn a bijection between images and some base random variable such as standard Gaussian \citep{dinh2016density, kingma2018glow, papamakarios2021normalize}. 
Notably, 
the invertibility of NFs guarantees that any image is assigned with a valid probability, and NFs have shown higher degrees of robustness and better performance than GANs, especially on reconstructions of out-of-distribution images  \citep{asim2020invertible, whang21solve, whang2021composing, li2021traversing, hong2022robustness}. In this paper, we focus on inverse problems incorporating an NF model as the generative prior. We give more details of NFs in Section~\ref{sec:literature}.

Conceptually, the aforementioned approaches can be seen as reconstructing an image as $\xv^*$ defined by
\begin{align}
\label{eqn:reconstruction}
    \xv^* (\lambda)\in \argmin\nolimits_{\xv} \mathcal{L}_{\text{recon}}(\xv, \yv) + \lambda  \mathcal{L}_{\text{reg}}(\xv),
\end{align}
where $\mathcal{L}_\text{recon}$ is a reconstruction error between the observation $\yv$ and a recovered image $\xv$~\citep{bora2017compressed,ulyanov2018deep}, and $\mathcal{L}_{\text{reg}}(\xv)$ multiplied by the hyperparmeter $\lambda$ regularizes the reconstruction $\xv$ using the prior information. Specifically, when a probabilistic deep generative prior, like an NF model, is used,  $\mathcal{L}_{\text{reg}}(\xv)$ can be the likelihood for the generative model to synthesize $\xv$.
Note that the loss function is subject to different noise models~\citep{van2018compressed,asim2020invertible,whang21solve}.


In execution, the reconstruction can be challenging as 
$\mathcal{L}_{\text{reg}}$ involves a deep generative prior. 
The success fundamentally relies on an effective optimization algorithm to find the global or a satisfactory local minimum of \eqref{eqn:reconstruction}. 
However, the non-convex nature of inverse problems often makes gradient descent unprincipled and non-robust, e.g., to initialization. In fact, even in a simpler problem where the forward operator is the identity map (corresponding to a denoising problem), solving \eqref{eqn:reconstruction} with a deep generative prior is NP-hard as demonstrated in \citet{lei2019inverting}. This establishes the complexity of solving inverse problems in general.  
On the other hand, even for specific cases, gradient descent possibly fails to find global optima, 
unlike training an (over-parameterized) neural network. This is 
because inverse problems require building a consistent (or under-parameterized)  system and yielding a unique solution. It is known both theoretically and empirically that the more over-parameterized the system is, the easier it is to find the global minima with first-order methods~\citep{jacot2018neural,du2019gradient,allen2019convergence}. 


In this paper, we overcome the difficulty by proposing a new principled optimization scheme for inverse problems with a deep generative prior. 
Our algorithm incrementally optimizes the reconstruction conditioning on a sequence of
$\lambda$'s that are gradually increased from 0 to a prespecified value. Intuitively, suppose we have found a satisfactory solution (e.g., the global optimum) $\xv^*(\lambda)$ as in \eqref{eqn:reconstruction}. Then with a small increase $\Delta\lambda$ in the hyperparameter, the new solution $\xv^*(\lambda+\Delta\lambda)$ should be close to $\xv^*(\lambda)$ and easy to find if starting from $\xv^*(\lambda)$.
Our algorithm is related to amortized optimization \citep{amos2022tutorial} in that the difficulty and the high computing cost of finding $\xv^*(\lambda)$ for the original inverse problem is amortized over a sequence of much easier tasks, where finding $\xv^*(0)$ is feasible and the solution to one task facilitates solving the next. We refer to our method as Amortized Inverse Problem Optimization (\notion). 

It is noteworthy that {\notion} is different from the amortized optimization in the conventional sense, which uses learning
to approximate/predict the solutions to similar problems \citep{amos2022tutorial}. In stark contrast, we are spreading the difficulty in solving the original problem into a sequence of much easier tasks, each of which is still the optimization of an inverse problem objective function. 
We provide a theoretical underpinning of {\notion}: 
Under some conventional assumptions, {\notion} is guaranteed to find the global minimum. A practical and efficient algorithm is also provided. Empirically, our algorithm exhibits superior performance in minimizing the loss of various inverse problems, including denoising, noisy compressed sensing, and inpainting. 
To the best of our knowledge, {\notion} is the first principled and efficient algorithm for solving inverse problems with a flow-based prior. 

The paper proceeds as follows. In Section~\ref{sec:literature}, we provide background knowledge in normalizing flows and amortized optimization and introduce example inverse problems. In Section~\ref{sec:method}, we formally propose {\notion} and give theoretical analysis. In Section~\ref{sec:experiment}, we illustrate our algorithm and show its outstanding performance compared to conventional methods that have $\lambda$ fixed during optimization. Section~\ref{sec:conclusion} concludes the paper. We defer  the proofs, some technical details and experiment settings, and supplementary  results to the appendix.


%% file: subfiles/2_related_work.tex
\section{Backgrounds}
\label{sec:literature}

We first provide an overview of normalizing flows that are used as the generative prior in our setting. We also briefly introduce amortized optimization based on learning and highlight its difference from the proposed {\notion}. In addition, we showcase three representative inverse problem tasks, on which our algorithm will be evaluated. 

\subsection{Normalizing Flows}

Normalizing flows (NFs) \citep{rezende2015variational, papamakarios2021normalize} are a family of generative models and  capable of representing an $n$-dimensional complex distribution by transforming it to a simple base distribution (e.g., standard Gaussian or uniform distribution) of the same dimension. Compared to other generative models such as GAN  \citep{goodfellow2014generative} and variational autoencoders  \citep{kingma2013auto}, NFs use a bijective (invertible) mapping and  are computationally flexible in the sense that they admit sampling from the distribution efficiently and conduct exact likelihood estimation. 

To be more specific, 
let $\xv \in \R^n$ denote a data point that follows an unknown complex distribution and $\zv \in \R^n$ follow some pre-specified based distribution such as a standard Gaussian.
An NF model learns a differentiable bijective function $G: \R^n \rightarrow \R^n$ such that $\xv = G(\zv)$. 
To sample from the data distribution $p_G(\xv)$, one can first generate $\zv \sim p(\zv)$ and then apply the transformation $\xv = G(\zv)$. Moreover, the invertibility of $G$ allows one to use the change-of-variable formula to calculate 
the likelihood of $\xv$ by
\begin{align*}
\log p_G(\xv) = \log p(\zv) + \log | \det (J_{G^{-1}}(\xv))|,  
\end{align*}
where $J_{G^{-1}}$ denotes the Jacobian matrix of the inverse mapping $G^{-1}$ evaluated at $\xv$. 
To speed up the computation, $G$ is usually composed of several simpler invertible functions that have triangular Jacobian matrices. 
Typical NFs include RealNVP \cite{dinh2016density} and GLOW \cite{kingma2018glow}. For more details of NF models, we refer the readers to the review by \citet{papamakarios2021normalize} and the references therein.

\subsection{Amortized Optimization Based on Learning}
Amortized optimization methods based on learning are used to improve repeated solutions to the  optimization problem
\begin{equation}\label{eq:amot_opt}
    \xv^\star(\lambdav)\in \argmin\nolimits_\xv g(\xv;\lambdav), 
\end{equation}
where the non-convex objective $g:\mathcal{X}\times\mathcal{A}\rightarrow \R$ takes some context $\lambdav\in\mathcal{A}$ that can be continuous or discrete. The continuous, unconstrained domain of the problem is given by $\xv\in\mathcal{X}=\R^n$, and the solution $\xv^\star (\lambdav)$, defined implicitly by the optimization process, is usually assumed to be unique. Given different $\lambdav$, instead of optimizing each $\xv^\star (\lambdav)$ separately, amortized optimization utilizes the similarities between subroutines induced by different $\lambdav$ to  amortize 
the computational difficulty and cost across them and gets its name thereof. 
Typically, 
an amortized optimization method \cite{amos2022tutorial} solving \eqref{eq:amot_opt} can be represented by
\begin{align}
\mathcal{M}\triangleq (g,\mathcal{X},\mathcal{A},p(\lambdav),\hat{x}_\theta,\mathcal{L}), 
\notag
\end{align}
where $g: \mathcal{X}\times \mathcal{A}\rightarrow \R$ is the unconstrained objective to optimize, $\mathcal{X}$ is the domain, $\mathcal{A}$ is the context space, $p(\lambdav)$ is the probability distribution over contexts to optimize, $\hat{x}_\theta: \mathcal{A}\rightarrow \mathcal{X}$ is the amortized model parameterized by $\theta$, which is {learned} by optimizing a loss $\mathcal{L}(g,\mathcal{X},\mathcal{A},p(\lambdav),\hat{x}_\theta)$ 
 defined on all the components \citep{kim2018semi,marino2018iterative,marino2021learned,liu2022teaching}.

Learning-based amortized optimization has been used in various machine learning models, including 
variational inference \citep{kingma2013auto},
model-agnostic meta-learning \citep{finn2017model}, multi-task learning \citep{stanley2009hypercube}, sparse coding \citep{chen2021learning},  reinforcement learning \citep{ichnowski2021accelerating}, and so on.
For a more comprehensive survey of amortized optimization, we refer the readers to \citet{amos2022tutorial}. Our proposed {\notion} is different from the learning-based amortized optimization in two aspects. First, we decompose the task of an inverse problem into easier ones and still require optimization, rather than learning, to solve each subroutine problem. Second,  the easier tasks in {\notion} are not independent; the solution to one task is used as the initial value for the next and facilitates its optimization.

\subsection{Representative Inverse Problems}
We briefly introduce three representative inverse problems that we use to validate {\notion} and refer the readers to \citet{ongie2020deep} for recent progress using deep learning. 
Given an unknown clean image $\xv_T$, we observe a corrupted measurement $\yv = f(\xv_T) + \ev$, 
where $f: \R^n \rightarrow \R^m$ is some known forward operator such that $m \leq n$.
The additive term $\ev \in \R^m$ denotes some random noise that 
is usually assumed to have independent and identically distributed entries \citep{bora2017compressed, asim2020invertible}.
{Representative inverse problem tasks} include denoising, noisy compressed sensing, inpainting, and so on, with  different forward operators $f$. In this work, we focus on the following three tasks.

\textbf{Denoising}
assumes that $\yv = \xv_T + \ev$ 
and noise $\ev \sim \gaus(0, \sigma^2\Imat)$ is an isotropic Gaussian vector \citep{asim2020invertible, ongie2020deep}.

\textbf{Noisy Compressed Sensing (NCS)}
assumes that $\yv = \Amat \xv_T + \ev$ where 
$\Amat \in \R^{m \times n}$, $m < n$, is a known $m \times n$ matrix of i.i.d. $\gaus(0, 1 / m)$ entries \citep{bora2017compressed, asim2020invertible, ongie2020deep},
and the noise $\ev \sim \gaus(0, \sigma^2\Imat)$ is an isotropic Gaussian vector. 
Typically, the smaller $m$ is, the more difficult the NCS task will be. 

\textbf{Inpainting}
assumes that $\yv = \Amat \xv_T + \ev$ where $\Amat \in \R^{n \times n}$ is a diagonal matrix with binary entries and a certain proportion of them are zeros. In other words, $\Amat$ indicates whether a pixel is observed or missing \citep{asim2020invertible, ongie2020deep}.
Again, we consider the noise $\ev \sim \gaus(0, \sigma^2\Imat)$.

%% file: subfiles/3_method.tex
\section{Methodology}
\label{sec:method}
We propose the Amortized  Inverse Problem Optimization (\notion) algorithm to reconstruct images by maximum a posterior estimation. First, we formulate the loss function for inverse problems using an NF generative prior.  Then we introduce the {\notion} algorithm and show the theoretical guarantee of its convergence.
\subsection{Maximum A Posterior Estimation}
Recent work on inverse problems using deep generative priors \citep{asim2020blind, whang21solve} has established successes in image reconstruction
by a maximum a posterior (MAP) estimation. We adopt the same MAP formulation as in \citet{whang21solve}. 
Specifically, we use a pre-trained invertible NF model $G: \R^n \rightarrow \R^n$ as the prior that we can effectively sample from.
$G$ maps the latent variable $\zv \in \R^n$ to an image $\xv \in \R^n$ and induces a tractable distribution over $\xv$ by the change-of-variable formula \citep{papamakarios2021normalize}. 
Optimization with respect to $\xv$ or $\zv$ has been considered in the literature \citep[e.g.,][]{asim2020invertible}.
In our context, they make no difference due to the invertibility of~$G$, and thus we directly optimize $\xv$ by minimizing the MAP loss. 
To be specific, denoting the prior density of $\xv$ as $p_G(\xv)$, which quantifies how likely an image $\xv$ is sampled from the pre-trained NF model, we reconstruct the image from $\yv$ by the MAP estimation 
\begin{align}
\xv^*(\lambda) \in &\argmin\nolimits_\xv \mathcal{L}_{\text{MAP}}(\xv; \lambda) \label{eq:map} \\
 = &\argmin\nolimits_\xv\  -\log p_e(\yv - f(\xv)) -\lambda \log p_G(\xv)\,, \nonumber
\end{align}
where the hyperparameter $\lambda > 0$ controls the weight of the prior, and 
$p_e$ is the density of the noise $\ev$. 
$-\log p_e(\yv - f(\xv))$ and $-\lambda \log p_G(\xv)$ are the reconstruction error and the regularization in \eqref{eqn:reconstruction}, respectively, both of which are continuous in $\xv$. In practice, $p_e$ is usually assumed to be an isotropic Gaussian distribution \citep{bora2017compressed, asim2020invertible, ongie2020deep} whose coordinates are independent and identically Gaussian distributed with a zero mean. 
Consequently, 
the loss function for the MAP estimation is equivalent to  
\begin{align}
    \mathcal{L}_{\text{MAP}}(\xv; \lambda) 
    = \| \yv - f(\xv) \|^2 -\lambda \log p_G(\xv).
    \notag
\end{align}
Note that the reconstruction error  reaches its minimum value if and only if $\yv = f(\xv)$.
It is challenging to directly minimize $\mathcal{L}_{\text{MAP}}(\xv; \lambda)$ given a prespecified $\lambda$ in the presence of the deep generative prior $p_G$ because of its non-convexity and NP-hardness \citep{lei2019inverting}. To effectively  and efficiently solve the problem, we propose to amortize the difficulty and computing cost over a sequence of easier subroutine optimization and provide theoretical guarantees. 

\subsection{Amortized Optimization for MAP}

We propose Amortized Inverse Problems Optimization ({\notion}) for solving \eqref{eq:map}. 
Given a prespecified hyperparameter value $\Lambda$, to obtain a good approximation of $\xv^*(\Lambda)$, we start from ${\lambda = 0}$
, where the optimization  may have an analytical solution ${\xv^*(0) \in\arg\min_{\xv} \loss_{\text{MAP}}(\xv; 0)= f^{-1}(\yv)}$, 
and gradually increase $\lambda$ towards $\Lambda$ in multiple steps.
In each step, assuming that the current solution $\xv^*(\lambda)$ is obtained and given a small enough $\Delta \lambda>0$, we expect $\xv^*(\lambda)$ to lie close to  
the solution  $\xv^*(\lambda + \Delta \lambda)$ in the next step under regular conditions (see Section~\ref{sec:map-fix-beta-thm}, shortly). 
In other words, $\xv^*(\lambda)$ is nearly optimal for minimizing the MAP loss  ${\loss_{\text{MAP}}(\xv; \lambda + \Delta \lambda)}$.
Consequently, minimizing ${\loss_{\text{MAP}}(\xv; \lambda + \Delta \lambda)}$ starting from $\xv^*(\lambda)$ makes the optimization easier and converge faster than starting from random initialization.
In particular, we amortize the difficulty in directly solving $\xv^*(\Lambda)$ over solving a sequence of optimization problems ${\{\min\nolimits_\xv\loss_{\text{MAP}}(\xv; \lambda_{i+1}=\lambda_i + \Delta \lambda_i)\given \xv^*(\lambda_i)\}_i}$. 

Notably, the starting point $\xv^*(0)$ corresponds to maximizing the log-likelihood of the noise $\ev$ and equals to the maximum likelihood estimation (MLE). However, not all inverse problems admit a unique MLE.  
For under-determined $f$, there exist infinitely many choices of $\xv^*(0)$ such that $f(\xv^*(0)) = \yv$ (e.g., in NCS and inpainting tasks), 
%
among which we choose the initial value of AIPO as the MLE $\xv^*(0)$ defined by
\begin{align}
\xv^* (0) \in \argmax\nolimits_\xv p_G(\xv), \ \text{s.t.} \ f(\xv) = \yv \label{eq:mle}.
\end{align}
In practice, \eqref{eq:mle} can be solved by projected gradient descent~\citep{boyd2004convex}. 
Specifically, all the tasks we consider in this paper have a linear forward operator $f$; the constraints are in the affine space, and the projection can be readily solved as presented in Appendix~\ref{app:projected}. 
We summarize {\notion} in Algorithm~\ref{alg:cascadex}. Note that its theoretical guarantee in Section~\ref{sec:map-fix-beta-thm}, shortly, does not rely on the linearity of $f$.

\begin{remark}
In denoising tasks, existing literature largely uses $\yv$ for initialization  
\citep{asim2020invertible, whang21solve} and can be regarded as a special case of our method by taking one large step from $\lambda=0$ with $\Delta \lambda = \Lambda$. 
\end{remark}




\begin{algorithm}[!t]
    \caption{{\notion} algorithm}
    \label{alg:cascadex}
    \begin{algorithmic}[1] 
        \STATE \textbf{Input:}  $\Lambda>0$, generative model $G:\R^n\to \R^n$, $L>0$ (see Assumption \ref{asmp:L_smooth_of_F}), $\sigma>0,\delta>0$ (see Assumption~\ref{asmp:local}),         $C>0$ (see Assumption \ref{asmp:C_smooth_of_x}), precision $\varepsilon>0$
        \STATE \textbf{Initialize:}
         $\lambda=0$, $\mu=\frac{1}{2L\sigma^2}$, $\delta_0=\min\left\{\delta,\frac{\mu}{\sqrt{2(L-\mu)L}}\delta\right\}$, $N=[\frac{2\Lambda C}{\delta_0}]+1$
        \STATE Find the MLE $\xv_0=\xv^*(0)$ by solving \eqref{eq:mle}
        \FOR{$i=0, \dots, N-1$} 
            \STATE $\lambda=\lambda+\frac{\Lambda}{N}$
            \STATE $K=[\frac{2\log(2\delta/\delta_0)}{\log(L/(L-\mu))}]+1$
            \IF{$i=N-1$}
             \STATE $K=\max\{0,\frac{2\log(2\delta/\varepsilon)}{\log(L/(L-\mu))}]+1\}$
            \ENDIF
            \FOR{$k=1, \dots, K$} 
            \STATE $\xv_{k+1}=\xv_k-\frac{1}{L}\nabla_{\xv}\mathcal{L}_\text{MAP}(\xv_k, \lambda)$
            \ENDFOR
            \STATE $\xv_0 = \xv_{K+1}$
        \ENDFOR 
        \OUTPUT $\hat\xv(\Lambda) = \xv_0$
    \end{algorithmic}
\end{algorithm}

%% file: subfiles/4_theorem.tex
\subsection{Theoretical Analysis}
\label{sec:map-fix-beta-thm}


We provide a theoretical analysis of the convergence of {\notion}. 
We make the following assumptions, under which we prove that {\notion} by Algorithm~\ref{alg:cascadex} finds an approximation of the global minimum of $\mathcal{L}_\text{MAP}$ with arbitrary precision.
\begin{assumption}[$L$-smoothness of $\mathcal{L}_\text{MAP}$]
\label{asmp:L_smooth_of_F}
There exists $L>0$ such that $\forall \lambda \in [0,\Lambda]$, $\mathcal{L}_\text{MAP}(\cdot; \lambda)$ is $L$-smooth, i.e., for all $\xv_1$ and $\xv_2$, $\norm{\nabla_\xv \mathcal{L}_\text{MAP}(\xv_1; \lambda)-\nabla_\xv \mathcal{L}_\text{MAP}(\xv_2; \lambda)}\leq L\norm{\xv_1-\xv_2}$.
\end{assumption}

\begin{assumption}[local property of $\nabla_\xv \mathcal{L}_\text{MAP}$]
\label{asmp:local}
There exists $ \sigma>0$ and $\delta>0$ such that for all $\lambda\in [0,\Lambda]$ and $\xv\in B(\xv^*(\lambda),\delta):=\{\xv \mid \norm{\xv-\xv^*(\lambda)}\leq\delta\}$, we have $\norm{\xv-\xv^*(\lambda)}\leq\sigma\norm{\nabla_\xv \mathcal{L}_\text{MAP}(\xv; \lambda)}$.
\end{assumption}
\begin{assumption}[$C$-smoothness of $ \xv^*(\lambda)$]
\label{asmp:C_smooth_of_x}
For all $\lambda\in(0,\Lambda]$, $\xv^*(\lambda)$ is unique, and 
there exists $C>0$ such that  for all $\lambda_1,\lambda_2\in (0,\Lambda]$, $\norm{\xv^*(\lambda_1)- \xv^*(\lambda_2)}\leq C|\lambda_1-\lambda_2|.$
\end{assumption}

\begin{remark}
\label{rmk:asmp_local}

Smoothness assumptions like Assumptions \ref{asmp:L_smooth_of_F} and \ref{asmp:C_smooth_of_x} are commonly used in convergence analysis. See, for example, \citet[][Assumption 3.2]{song2019surfing}, \citet[][Assumption 1]{zhou2019sgd}, and \citet{scaman2022convergence}.

We make two comments on Assumption \ref{asmp:local}: 

(i) Local strong convexity around the minima of the loss is a widely-adopted assumption in deep learning literature,  e.g., \citet[][Definition 2.4]{li2017convergence}, \citet[][Theorem 4.1]{whang2020compressed}, and \citet[][Definition 5]{safran2021effects}.
Our Assumption~\ref{asmp:local} is weaker than the local strong convexity. Reversely, if there exists $\mu'>0$ and $\delta>0$ such that for all $\lambda\in[0,\Lambda]$, $\mathcal{L}_\text{MAP}(\cdot; \lambda)$ is $\mu'$-strongly convex on $B(\xv^*(\lambda),\delta)$, then Assumption \ref{asmp:local} holds with $\sigma=2/\mu'$. More Details about this comment can be found in Appendix~\ref{sec_app:pf_rmk}.

(ii) If Assumption~\ref{asmp:L_smooth_of_F} holds, then Assumption \ref{asmp:local} implies the local Polyak-Lojasiewicz condition. That is to say, if Assumptions~\ref{asmp:L_smooth_of_F} and~\ref{asmp:local} hold, then for all $\lambda \in [0,\Lambda]$ and $\xv\in B(\xv^*(\lambda),\delta)$, we have
\begin{equation*} \label{eq:2}
    \mathcal{L}_\text{MAP}(\xv; \lambda)-\mathcal{L}_\text{MAP}(\xv^*(\lambda); \lambda)\leq\frac{1}{2\mu}\norm{\nabla_\xv \mathcal{L}_\text{MAP}(\xv; \lambda)}^2\,,
\end{equation*}
where $\textstyle{\mu=\frac{1}{2L\sigma^2}}$. More details about this comment can be found in Appendix.~\ref{sec_app:pf_rmk}.

The local Polyak-Lojasiewicz property is previously used to characterize the local optimization landscape for training neural networks 
\cite{song2021subquadratic, karimi2016linear, liu2022loss}. It has been shown that wide neural networks satisfy the local Polyak-Lojasiewicz property under mild assumptions \citep[][Theorem 7.2]{liu2020toward}.
\end{remark}
Now we present our main result.
\begin{theorem} \label{thm:cascade}
Under Assumptions \ref{asmp:L_smooth_of_F}, \ref{asmp:local}, and \ref{asmp:C_smooth_of_x}, for all $\varepsilon>0$, Algorithm \ref{alg:cascadex} returns $\hat\xv(\Lambda)$ that satisfies $\norm{\hat\xv(\Lambda)-\xv^*(\Lambda)}\leq \varepsilon$.
\end{theorem}
Note that Theorem~\ref{thm:cascade} ensures that Algorithm~\ref{alg:cascadex} for {\notion} finds an $\varepsilon$-approximate point of the \textbf{global} minimum of $\mathcal{L}_\text{MAP}(\cdot; \Lambda)$.
We give a proof sketch of Theorem~\ref{thm:cascade} here and defer the formal proof to the appendix. 
\begin{proof}[Proof sketch]
Starting from the global minimum when ${\lambda=0}$, 
our algorithm ensures that the $i$-th outer iteration learns $\xv^*(i\Lambda/N)$ approximately and serves as a good initialization for the next target $\xv^*((i+1)\Lambda/N)$. Note that in each iteration we incrementally grow $\lambda$ from $i\Lambda/N$ to $(i+1)\Lambda/N$ until reaching our target $\Lambda$. Specifically, Assumptions~\ref{asmp:C_smooth_of_x} and~\ref{asmp:local}, respectively, ensure $\xv^*(i\Lambda/N)$ to be close enough to $\xv^*((i+1)\Lambda/N)$ and that the first order algorithm can find $\xv^*((i+1)\Lambda/N)$ from the good initialization  obtained in the last iteration. 
\end{proof}

To avoid specifying the parameters in the assumptions ($L,\sigma,\delta,C$) and the precision $\varepsilon$, we  provide an efficient and practical implementation of {\notion} in Algorithm~\ref{alg:cascadex2} in the appendix, where the scheme for hyperparameter increment is data-adaptive.

%% file: subfiles/5_experiment.tex
\section{Experiments}
\label{sec:experiment}



In this section, we evaluate the performance of the proposed algorithm on three inverse problem tasks, including denoising, noisy compressed sensing, and inpainting. We use two normalizing flow models as the generative prior, both of which work well, to justify that our algorithm is a general framework and does not require model-specific adaption.
Note that using deep generative priors in inverse problems has been demonstrated to outperform classical approaches \citep{bora2017compressed, asim2020invertible, whang21solve}. 
We focus on illustrating the advantage of {\notion} over conventional optimizations without the amortization scheme and skip the comparison with classical approaches. 

\subsection{Setup}

We use two commonly used normalizing flow models, RealNVP \citep{dinh2016density} and GLOW \citep{kingma2018glow}, respectively, as the generative prior $G$. 
The two models  are trained on the CelebA dataset \cite{liu2015faceattributes}, and we follow the pertaining suggestions by \citet{asim2020invertible} and \citet{whang21solve},  to which we refer the readers for the model architecture and technical details.

Our experiments consist of two sets of 
data. One is in-distribution samples that are randomly selected from the CelebA test set. The other is out-of-distribution (OOD) images that contain human or human-like objects.
Due to budget constraints, we run the experiments on 200 in-distribution and 7 OOD samples. 
For baseline algorithms, we consider minimizing the MAP loss as in equation \eqref{eq:map} by  gradient descent with random or zero initialization that is widely used in literature \cite{bora2017compressed, asim2020invertible, whang21solve}. Concretely,
random initialization first draws a Gaussian random vector $\zv_0$ and uses $\xv_0 = G(\zv_0)$ as the initial value. Zero initialization takes $\zv_0 = \zeros$ and initializes $\xv_0 = G(\zv_0)$. 
Furthermore, to demonstrate that {\notion}'s outperformance indeed results from amortization rather than merely the MLE initialization, gradient descent with the MLE initialization is used as the third baseline approach. 

All the three baselines have $\lambda$ fixed throughout the optimization process.
In all the experiments, we optimize $\xv$ by Adam \citep{kingma2014adam} and assign an equal computing budget to all the approaches compared in each subsection. 
Our {\notion} and the baseline algorithm with the MLE initialization require a solution to \eqref{eq:mle} on the NCS and inpainting tasks, where we run 500 iterations of projected gradient descent. In these cases, we assign an extra computing budget of the same amount to the baseline algorithms with the
random and zero initialization.

In all the experiments, we compare the algorithms with a prespecified $\lambda$, which is set to be $0.3,0.5,1.0,1.5,2.0$, respectively. These values form a fairly large range for the hyperparameter \cite{asim2020invertible, whang21solve}.
When evaluating an algorithm's performance, we rely on two metrics: the MAP loss that the algorithms try to minimize as defined in equation~\eqref{eq:map} and Peak-Signal-to-Noise Ratio (PSNR) whose large value indicates a better reconstruction. We also show the reconstructed images for illustration. Due to the page limit, we only report the MAP loss values and visualize reconstructions using one generative prior per task. We defer PSNRs and more visualization to the appendix.



\begin{table}[tb]
    \begin{center}
    \caption{
    MAP loss (mean$\pm$se) from solving denoising tasks with different algorithms under two generative priors. Lower is better.
    }
    \label{tab:denoise-loss}
    \resizebox{\linewidth}{!}{
    \renewcommand{\tabcolsep}{6pt}
    \begin{tabular}{rrrrrrr}
    \toprule
    & & $\lambda$ & Ours. & MLE init. & Random init.  & Zero init. \\
    \midrule
    \multirow{10}{*}{\rotatebox[origin=c]{90}{RealNVP}}
    &
    \multirow{5}{*}{\rotatebox[origin=c]{90}{Test}}
    &
    0.3 & \textbf{-9956 $\pm$ 19} & -9732 $\pm$ 20 & -9735 $\pm$ 21 & -9729 $\pm$ 20\\
    & & 
    0.5 & \textbf{-8909 $\pm$ 30} & -8376 $\pm$ 32 & -8357 $\pm$ 33 & -8385 $\pm$ 33\\
    & & 
    1.0 & \textbf{-6893 $\pm$ 61} & -5530 $\pm$ 65 & -5569 $\pm$ 65 & -5527 $\pm$ 67\\    
    & & 
    1.5 & \textbf{-4830 $\pm$ 95} & -2803 $\pm$ 105 & -2850 $\pm$ 106 & -2571 $\pm$ 103\\
    & & 
    2.0 & \textbf{-2619 $\pm$ 148} & -266 $\pm$ 134 & -330 $\pm$ 129 & 415 $\pm$ 143\\
    \cmidrule{3-7}
    &
    \multirow{5}{*}{\rotatebox[origin=c]{90}{OOD}}
    &
    0.3 & \textbf{-9680 $\pm$ 49} & -9464 $\pm$ 93 & -9451 $\pm$ 91 & -9513 $\pm$ 94\\
    & & 
    0.5 & \textbf{-8713 $\pm$ 110} & -7917 $\pm$ 262 & -7924 $\pm$ 191 & -7862 $\pm$ 242\\
    & & 
    1.0 & \textbf{-6444 $\pm$ 429} & -4413 $\pm$ 566 & -4610 $\pm$ 442 & -4751 $\pm$ 559\\
    & & 
    1.5 & \textbf{-4708 $\pm$ 920} & -1725 $\pm$ 829 & -1287 $\pm$ 793 & -1611 $\pm$ 760\\
    & & 
    2.0 & \textbf{-1669 $\pm$ 1567} & 1320 $\pm$ 970 & 1794 $\pm$ 857 & 1436 $\pm$ 949\\
    \cmidrule{2-7}

    \multirow{10}{*}{\rotatebox[origin=c]{90}{GLOW}}
    &
    \multirow{5}{*}{\rotatebox[origin=c]{90}{{Test}}}
    &
    0.3 & \textbf{-10368 $\pm$ 20} & -10286 $\pm$ 19 & -10293 $\pm$ 20 & -10294 $\pm$ 19\\    
    & & 
    0.5 & \textbf{-9902 $\pm$ 34} & -9657 $\pm$ 33 & -9646 $\pm$ 33 & -9626 $\pm$ 33\\
    & & 
    1.0 & \textbf{-9232 $\pm$ 62} & -8653 $\pm$ 57 & -8614 $\pm$ 58 & -8646 $\pm$ 59\\
    & & 
    1.5 & \textbf{-8893 $\pm$ 87} & -8011 $\pm$ 78 & -7940 $\pm$ 79 & -7912 $\pm$ 80\\
    & & 
    2.0 & \textbf{-8851 $\pm$ 107} & -7485 $\pm$ 96 & -7461 $\pm$ 98 & -7409 $\pm$ 108\\
    \cmidrule{3-7}
    &
    \multirow{5}{*}{\rotatebox[origin=c]{90}{OOD}}
    &
    0.3 & \textbf{-10329 $\pm$ 181} & -10187 $\pm$ 211 & -10266 $\pm$ 182 & -10242 $\pm$ 177\\
    & & 
    0.5 & \textbf{-9843 $\pm$ 331} & -9495 $\pm$ 308 & -9421 $\pm$ 198 & -9465 $\pm$ 238\\
    & & 
    1.0 & \textbf{-9122 $\pm$ 454} & -8499 $\pm$ 464 & -8721 $\pm$ 379 & -8468 $\pm$ 410\\
    & & 
    1.5 & \textbf{-9254 $\pm$ 584} & -7841 $\pm$ 598 & -8005 $\pm$ 540 & -8072 $\pm$ 492\\
    & & 
    2.0 & \textbf{-9409 $\pm$ 756} & -7673 $\pm$ 640 & -6978 $\pm$ 727 & -7393 $\pm$ 583\\    
    \bottomrule
    \end{tabular}
    }
    \end{center}
\end{table}

\begin{figure}[tb]
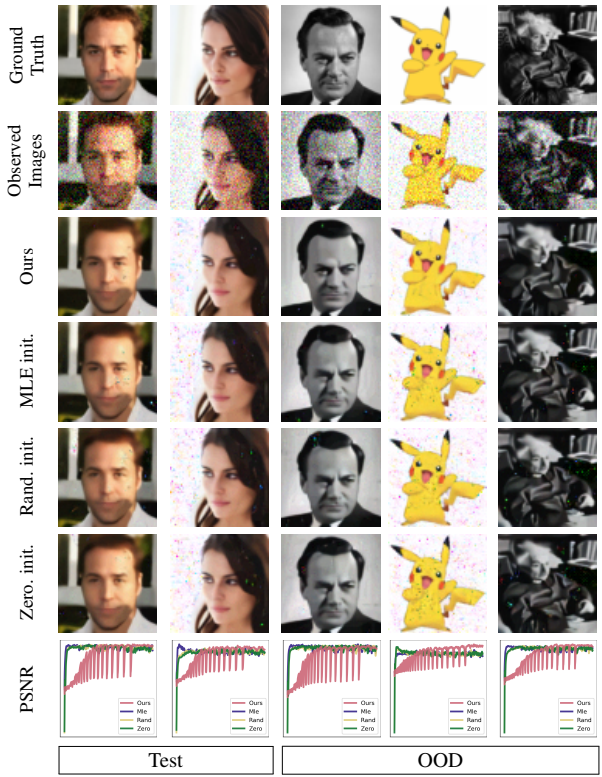

    \begin{center}

    \resizebox{\linewidth}{!}{
    \renewcommand{\tabcolsep}{2pt}
    \def\figwidth{0.2\linewidth}%
    \newcommand{\authornote}[1]{
        \footnotesize 
        \adjustbox{rotate=90}{\parbox{\figwidth}{\centering #1}}
    }
    \begin{tabular}{*{11}{c}}
      \authornote{Ground Truth}     &  
        \loadimage{figures/main/denoise/realnvp}{width=\figwidth}{0}{5}\\
      \authornote{Observed Images}     & 
        \loadimage{figures/main/denoise/realnvp}{width=\figwidth}{10}{5}\\
      \authornote{Ours}        & 
        \loadimage{figures/main/denoise/realnvp}{width=\figwidth}{20}{5}\\
      \authornote{MLE init.}        & 
        \loadimage{figures/main/denoise/realnvp}{width=\figwidth}{30}{5}\\
      \authornote{Rand. init.}          &  
        \loadimage{figures/main/denoise/realnvp}{width=\figwidth}{40}{5}\\
      \authornote{Zero. init.}         &  
        \loadimage{figures/main/denoise/realnvp}{width=\figwidth}{50}{5}\\
      \authornote{PSNR}         &  
        \loadimage{figures/main/denoise/realnvp}{width=\figwidth}{60}{5}\\
      & \multicolumn{2}{c}{\fbox{\parbox{\calctotalwidth{\figwidth}{2}}{\centering Test}}} 
      & \multicolumn{3}{c}{\fbox{\parbox{\calctotalwidth{\figwidth}{3}}{\centering OOD}}}
    \end{tabular}
    }
    \caption{
    Reconstructions from denoising Gaussian noise on CelebA faces and out-of-distribution images with RealNVP as the generative prior. 
    Hyperparameter $\lambda = 0.5$.
    }
    \label{fig:denoise-realnvp}
    \end{center}
\end{figure}

\subsection{Denoising}
Denoising tasks assume that $\yv=\xv + \ev$, and we sample $\ev \sim \gaus(0, 0.1^2\Imat)$ \citep{bora2017compressed, asim2020invertible}.
We report the corresponding MAP losses in Table \ref{tab:denoise-loss}. As a result, {\notion}  consistently outperforms the three baselines regardless of which generative prior is used. In particular, when $\lambda$ is set to be large ($1.5$ or $2.0$), the \notion's outperformance is especially remarkable. This is primarily because the non-convex $p_G$ weighs more in the MAP loss with a larger $\lambda$, increasing the difficulty in conventional optimization without the amortization. In stark contrast, {\notion} is less sensitive to $\lambda$ as it amortizes the difficulty over much simpler tasks. Consequently, given the same budget, {\notion} delivers much smaller loss values. It is also the case in the NCS and inpainting tasks as shown in the following subsections.

We visualize the reconstruction quality by showing the clean, observed, and reconstructed images. 
For this task, we report results from RealNVP in Figure \ref{fig:denoise-realnvp}. We set the hyperparameter $\lambda = 0.5$ such that all four algorithms achieve their highest PSNRs (see Tables~\ref{tab:realnvp-psnr} and~\ref{tab:glow-psnr} in the appendix for details). We also present the full version of Figure \ref{fig:denoise-realnvp}, reconstructions from GLOW, and their corresponding PSNRs in the appendix. 
Our method preserves the details of the images and removes more noise in the background. See, for instance, the second and the fourth columns in Figure \ref{fig:denoise-realnvp}. The reconstructions together with the consistently lower loss values demonstrate the success of our proposed approach.

\begin{table}[tb]
    \begin{center}
    \caption{
    MAP loss (mean$\pm$se) of solving NCS tasks with different algorithms under two generative priors. A smaller value is better.
    }
    \label{tab:csense-loss}
    \resizebox{\linewidth}{!}{
    \renewcommand{\tabcolsep}{6pt}
    \begin{tabular}{rrrrrrr}
    \toprule
    & & $\lambda$ & Ours. & MLE init. & Random init.  & Zero init. \\
    \midrule
    \multicolumn{7}{c}{ $m=4000$} \\
    \midrule
    \multirow{10}{*}{\rotatebox[origin=c]{90}{RealNVP}}
    &
    \multirow{5}{*}{\rotatebox[origin=c]{90}{{Test}}}
    &
    0.3 & \textbf{-2650 $\pm$ 19} & -2316 $\pm$ 20 & -2315 $\pm$ 19 & -2321 $\pm$ 19\\
    & & 
    0.5 & \textbf{-1686 $\pm$ 30} & -1084 $\pm$ 32 & -1006 $\pm$ 33 & -1024 $\pm$ 33\\
    & & 
    1.0 & \textbf{363 $\pm$ 61} & 1814 $\pm$ 64 & 1915 $\pm$ 63 & 1878 $\pm$ 65\\
    & & 
    1.5 & \textbf{2310 $\pm$ 97} & 4377 $\pm$ 98 & 4532 $\pm$ 100 & 5031 $\pm$ 96\\
    & & 
    2.0 & \textbf{4336 $\pm$ 126} & 6756 $\pm$ 124 & 7296 $\pm$ 128 & 7237 $\pm$ 128\\
    \cmidrule{3-7}
    &
    \multirow{5}{*}{\rotatebox[origin=c]{90}{{OOD}}}
    &
    0.3 & \textbf{-2433 $\pm$ 84} & -2088 $\pm$ 148 & -2058 $\pm$ 108 & -2043 $\pm$ 123\\
    & & 
    0.5 & \textbf{-1188 $\pm$ 152} & -604 $\pm$ 256 & -679 $\pm$ 222 & -598 $\pm$ 282\\
    & & 
    1.0 &\textbf{ 818 $\pm$ 604} & 2599 $\pm$ 611 & 2806 $\pm$ 615 & 2040 $\pm$ 482\\
    & & 
    1.5 & \textbf{3969 $\pm$ 1297} & 6052 $\pm$ 1011 & 5437 $\pm$ 922 & 5708 $\pm$ 762\\
    & & 
    2.0 & \textbf{6275 $\pm$ 1485} & 7830 $\pm$ 1252 & 9120 $\pm$ 1064 & 8645 $\pm$ 1044\\
    \cmidrule{2-7}

    \multirow{10}{*}{\rotatebox[origin=c]{90}{GLOW}}
    &
    \multirow{5}{*}{\rotatebox[origin=c]{90}{{Test}}}
    &
    0.3 & \textbf{-3110 $\pm$ 19} & -2915 $\pm$ 19 & -2901 $\pm$ 18 & -2895 $\pm$ 18\\
    & & 
    0.5 & \textbf{-2619 $\pm$ 33} & -2355 $\pm$ 31 & -2358 $\pm$ 31 & -2364 $\pm$ 31\\
    & & 
    1.0 & \textbf{-1949 $\pm$ 62} & -1350 $\pm$ 56 & -1392 $\pm$ 56 & -1392 $\pm$ 56\\
    & & 
    1.5 & \textbf{-1662 $\pm$ 85} & -693 $\pm$ 81 & -728 $\pm$ 79 & -713 $\pm$ 80\\
    & & 
    2.0 & \textbf{-1546 $\pm$ 105} & -151 $\pm$ 98 & -158 $\pm$ 96 & -58 $\pm$ 93\\
    \cmidrule{3-7}
    &
    \multirow{5}{*}{\rotatebox[origin=c]{90}{{OOD}}}
    &
    0.3 & \textbf{-3076 $\pm$ 201} & -2864 $\pm$ 185 & -2846 $\pm$ 172 & -2825 $\pm$ 191\\
    & & 
    0.5 & \textbf{-2647 $\pm$ 319} & -2306 $\pm$ 284 & -2252 $\pm$ 257 & -2349 $\pm$ 277\\
    & & 
    1.0 & \textbf{-2036 $\pm$ 457} & -1202 $\pm$ 475 & -1506 $\pm$ 511 & -1465 $\pm$ 452\\
    & & 
    1.5 & \textbf{-1563 $\pm$ 676} & -783 $\pm$ 687 & -869 $\pm$ 534 & -997 $\pm$ 588\\
    & & 
    2.0 & \textbf{-1610 $\pm$ 782} & -62 $\pm$ 758 & -322 $\pm$ 740 & -513 $\pm$ 650\\
    \midrule

    \multicolumn{7}{c}{$m=2000$} \\

    \midrule
    \multirow{10}{*}{\rotatebox[origin=c]{90}{RealNVP}}
    &
    \multirow{5}{*}{\rotatebox[origin=c]{90}{{Test}}}
    &
    0.3 & \textbf{-841 $\pm$ 18} & -498 $\pm$ 20 & -486 $\pm$ 19 & -465 $\pm$ 19\\
    & & 
    0.5 & \textbf{43 $\pm$ 31} & 733 $\pm$ 32 & 816 $\pm$ 32 & 751 $\pm$ 33\\
    & & 
    1.0 & \textbf{2027 $\pm$ 65} & 3436 $\pm$ 65 & 3615 $\pm$ 65 & 3679 $\pm$ 68\\
    & & 
    1.5 & \textbf{3953 $\pm$ 99} & 6124 $\pm$ 97 & 6349 $\pm$ 96 & 6408 $\pm$ 111\\
    & & 
    2.0 & \textbf{6422 $\pm$ 131} & 8484 $\pm$ 146 & 9076 $\pm$ 145 & 8761 $\pm$ 136\\
    \cmidrule{3-7}
    &
    \multirow{5}{*}{\rotatebox[origin=c]{90}{{OOD}}}
    &
    0.3 & \textbf{-566 $\pm$ 132} & -227 $\pm$ 146 & -203 $\pm$ 122 & -207 $\pm$ 146\\
    & & 
    0.5 & \textbf{184 $\pm$ 178} & 1295 $\pm$ 300 & 1158 $\pm$ 278 & 1385 $\pm$ 327\\
    & & 
    1.0 & \textbf{3062 $\pm$ 779} & 4341 $\pm$ 763 & 4624 $\pm$ 694 & 4169 $\pm$ 627\\
    & & 
    1.5 & \textbf{5100 $\pm$ 1063} & 7660 $\pm$ 1002 & 7021 $\pm$ 1021 & 7072 $\pm$ 748\\
    & & 
    2.0 & \textbf{9478 $\pm$ 1590} & 10774 $\pm$ 1683 & 10142 $\pm$ 1529 & 10315 $\pm$ 1235\\
    \cmidrule{2-7}

    \multirow{10}{*}{\rotatebox[origin=c]{90}{GLOW}}
    &
    \multirow{5}{*}{\rotatebox[origin=c]{90}{{Test}}}
    &
    0.3 & \textbf{-1378 $\pm$ 20} & -867 $\pm$ 28 & -986 $\pm$ 23 & -958 $\pm$ 24\\
    & & 
    0.5 & \textbf{-922 $\pm$ 32} & -295 $\pm$ 39 & -402 $\pm$ 36 & -399 $\pm$ 36\\
    & & 
    1.0 & \textbf{-355 $\pm$ 56} & 599 $\pm$ 59 & 412 $\pm$ 54 & 440 $\pm$ 55\\
    & & 
    1.5 & \textbf{-117 $\pm$ 85} & 1165 $\pm$ 77 & 1168 $\pm$ 77 & 1183 $\pm$ 76\\
    & & 
    2.0 & \textbf{-62 $\pm$ 103} & 1812 $\pm$ 105 & 1743 $\pm$ 102 & 1756 $\pm$ 102\\
    \cmidrule{3-7}
    &
    \multirow{5}{*}{\rotatebox[origin=c]{90}{{OOD}}}
    &
    0.3 & \textbf{-1323 $\pm$ 188} & -654 $\pm$ 174 & -660 $\pm$ 313 & -636 $\pm$ 322\\
    & & 
    0.5 & \textbf{-952 $\pm$ 300} & -356 $\pm$ 316 & -292 $\pm$ 216 & -274 $\pm$ 280\\
    & & 
    1.0 & \textbf{-25 $\pm$ 529} & 688 $\pm$ 604 & 2501 $\pm$ 1841 & 2461 $\pm$ 1833\\
    & & 
    1.5 & \textbf{-248 $\pm$ 663} & 1446 $\pm$ 756 & 938 $\pm$ 561 & 873 $\pm$ 536\\
    & & 
    2.0 & \textbf{116 $\pm$ 875} & 2200 $\pm$ 709 & 1605 $\pm$ 840 & 2014 $\pm$ 1023\\
    \bottomrule
    \end{tabular}
    }
    \end{center}
\end{table}

\begin{figure}[!t]
    \begin{center}
    \resizebox{\linewidth}{!}{
    \renewcommand{\tabcolsep}{2pt}
    \def\figwidth{0.2\linewidth}%
    \newcommand{\authornote}[1]{
        \footnotesize 
        \adjustbox{rotate=90}{\parbox{\figwidth}{\centering #1}}
    }
    \begin{tabular}{*{11}{c}}
      \authornote{Ground Truth}     &  
        \loadimage{figures/main/csense4000/glow}{width=\figwidth}{0}{5}\\
      \authornote{Ours}        & 
        \loadimage{figures/main/csense4000/glow}{width=\figwidth}{20}{5}\\
      \authornote{MLE init.}        & 
        \loadimage{figures/main/csense4000/glow}{width=\figwidth}{30}{5}\\
      \authornote{Rand. init.}          &  
        \loadimage{figures/main/csense4000/glow}{width=\figwidth}{40}{5}\\
      \authornote{Zero. init.}         &  
        \loadimage{figures/main/csense4000/glow}{width=\figwidth}{50}{5}\\
      \authornote{PSNR}         &  
        \loadimage{figures/main/csense4000/glow}{width=\figwidth}{60}{5}\\
      & \multicolumn{2}{c}{\fbox{\parbox{\calctotalwidth{\figwidth}{2}}{\centering Test}}} 
      & \multicolumn{3}{c}{\fbox{\parbox{\calctotalwidth{\figwidth}{3}}{\centering OOD}}}
    \end{tabular}
    }
    \caption{
    Reconstructions from NCS when $m=4000$ on CelebA faces and out-of-distribution images with GLOW as the generative prior. 
    Hyperparameter $\lambda=0.5$.
    }
    \label{fig:csense-glow-4000}
    \end{center}
\end{figure}

\begin{figure}[!ht]
    \begin{center}
    \resizebox{\linewidth}{!}{
    \renewcommand{\tabcolsep}{2pt}
    \def\figwidth{0.2\linewidth}%
    \newcommand{\authornote}[1]{
        \footnotesize 
        \adjustbox{rotate=90}{\parbox{\figwidth}{\centering #1}}
    }
    \begin{tabular}{*{11}{c}}
      \authornote{Ground Truth}     &  
        \loadimage{figures/main/csense2000/realnvp}{width=\figwidth}{0}{5}\\
      \authornote{Ours}        & 
        \loadimage{figures/main/csense2000/realnvp}{width=\figwidth}{20}{5}\\
      \authornote{MLE init.}        & 
        \loadimage{figures/main/csense2000/realnvp}{width=\figwidth}{30}{5}\\
      \authornote{Rand. init.}          &  
        \loadimage{figures/main/csense2000/realnvp}{width=\figwidth}{40}{5}\\
      \authornote{Zero. init.}         &  
        \loadimage{figures/main/csense2000/realnvp}{width=\figwidth}{50}{5}\\
      \authornote{PSNR}         &  
        \loadimage{figures/main/csense2000/realnvp}{width=\figwidth}{60}{5}\\
      & \multicolumn{2}{c}{\fbox{\parbox{\calctotalwidth{\figwidth}{2}}{\centering Test}}} 
      & \multicolumn{3}{c}{\fbox{\parbox{\calctotalwidth{\figwidth}{3}}{\centering OOD}}}
    \end{tabular}
    }
    \caption{
    Reconstructions from NCS when $m=2000$ on CelebA faces and out-of-distribution images with RealNVP as the generative prior. Hyperparameter $\lambda=0.3$.
    }
    \label{fig:csense-realnvp-2000}
    \end{center}
\end{figure}

\subsection{Noisy Compressed Sensing (NCS)}

Our second task is NCS in the form of $\yv = \Amat \xv + \ev$, where 
$\Amat \in \R^{m \times n}$ is a known matrix of i.i.d. Gaussian entries with $m < n=12288$.
The smaller $m$ is, the more difficult the NCS task will be. We consider $m = 2000$ and $4000$, respectively. 
Previous works~\citep{bora2017compressed, asim2020invertible} sample $\ev$ from Gaussian with the restrictions $\E[\ev] = \zeros$ and $\E [\| \ev \|] = 0.1$.
In this setting, however, the scale of the noise is not big enough to break down the algorithms in our experiments such that their performances are comparable.
So in our experiment, we let $\ev \sim \gaus(\zeros, 0.1^2\Imat)$, inducing larger noise and a more challenging NCS task. 

We report the MAP loss values from different algorithms in Table \ref{tab:csense-loss}. For all the NCS tasks in the setting of different generative priors and values of $m$ and $\lambda$, our proposed {\notion} consistently delivers much lower loss values compared to the benchmarks.
Analogous to the denoising tasks,  {\notion} outperforms the other algorithms by a larger margin in the presence of a bigger $\lambda$. 

We visualize the image reconstructions using the GLOW prior for $m=4000$ in Figure~\ref{fig:csense-glow-4000} and using the RealNVP prior for $m=2000$  in Figure~\ref{fig:csense-realnvp-2000}. The corresponding hyperparameters are 0.5 and 0.3, respectively. Unlike denoising and inpainting (in the next subsection), NCS does not have conceptually meaningful corrupted observations. So we omit to visualize them. 
Our algorithm is able to achieve better reconstruction qualities than the baselines, in that much less noise occurs in the background of the reconstructed images.
Moreover, compared to random and zero initialization, the MLE initialization does not exhibit consistently lower losses or consistently better reconstructions. This implies that the success of our {\notion} does not solely rely on the MLE initialization. Instead, the amortized optimization scheme contributes more to the improvement. 


\subsection{Inpainting}

\begin{table}[tb]
    \begin{center}
    \caption{
    MAP loss (mean$\pm$se) of solving inpainting tasks with different algorithms under two generative priors, lower is better.
    }
    \label{tab:inpaint-loss}
    \resizebox{\linewidth}{!}{
    \renewcommand{\tabcolsep}{6pt}
    \begin{tabular}{llrrrrr}

    \toprule
    & & $\lambda$ & Ours. & MLE init. & Random init.  & Zero init. \\
    \midrule
    \multirow{10}{*}{\rotatebox[origin=c]{90}{RealNVP}}
    &
    \multirow{5}{*}{\rotatebox[origin=c]{90}{{Test}}}
    &
    0.3 & \textbf{-31668 $\pm$ 16} & -31567 $\pm$ 17 & -31540 $\pm$ 18 & -31549 $\pm$ 18\\
    & & 
    0.5 &\textbf{ -29734 $\pm$ 28} & -29577 $\pm$ 31 & -29545 $\pm$ 30 & -29504 $\pm$ 30\\
    & & 
    1.0 & \textbf{-25940 $\pm$ 63} & -25309 $\pm$ 64 & -25241 $\pm$ 67 & -25304 $\pm$ 68\\
    & & 
    1.5 &\textbf{ -22871 $\pm$ 104} & -21565 $\pm$ 104 & -21305 $\pm$ 104 & -21450 $\pm$ 103\\
    & & 
    2.0 & \textbf{-20312 $\pm$ 139} & -17784 $\pm$ 149 & -18064 $\pm$ 146 & -17832 $\pm$ 148\\
    \cmidrule{3-7}
    &
    \multirow{5}{*}{\rotatebox[origin=c]{90}{{OOD}}}
    &
    0.3 & \textbf{-31295 $\pm$ 135} & -31110 $\pm$ 212 & -31123 $\pm$ 187 & -31231 $\pm$ 163\\
    & & 
    0.5 & \textbf{-29251 $\pm$ 220} & -28999 $\pm$ 250 & -28724 $\pm$ 315 & -28900 $\pm$ 283\\
    & & 
    1.0 & \textbf{-24963 $\pm$ 464} & -24581 $\pm$ 470 & -24179 $\pm$ 481 & -24014 $\pm$ 513\\
    & & 
    1.5 & \textbf{-21454 $\pm$ 715} & -20201 $\pm$ 678 & -19557 $\pm$ 836 & -19977 $\pm$ 850\\
    & & 
    2.0 & \textbf{-18418 $\pm$ 998} & -16029 $\pm$ 1082 & -15010 $\pm$ 1141 & -15501 $\pm$ 1039\\
    \cmidrule{2-7}
    
    \multirow{10}{*}{\rotatebox[origin=c]{90}{GLOW}}
    &
    \multirow{5}{*}{\rotatebox[origin=c]{90}{{Test}}}
    &
    0.3 & \textbf{-32037 $\pm$ 17} & -31964 $\pm$ 17 & -31968 $\pm$ 17 & -31966 $\pm$ 17\\
    & & 
    0.5 & \textbf{-30432 $\pm$ 33} & -30302 $\pm$ 32 & -30317 $\pm$ 32 & -30312 $\pm$ 32\\
    & & 
    1.0 & \textbf{-27518 $\pm$ 76} & -27158 $\pm$ 73 & -27173 $\pm$ 74 & -27164 $\pm$ 74\\
    & & 
    1.5 & \textbf{-25385 $\pm$ 122} & -24722 $\pm$ 115 & -24741 $\pm$ 115 & -24761 $\pm$ 112\\
    & & 
    2.0 & \textbf{-23726 $\pm$ 165} & -22670 $\pm$ 152 & -22660 $\pm$ 158 & -22656 $\pm$ 154\\
    \cmidrule{3-7}
    &
    \multirow{5}{*}{\rotatebox[origin=c]{90}{{OOD}}}
    &
    0.3 & \textbf{-31820 $\pm$ 146} & -31629 $\pm$ 149 & -31654 $\pm$ 159 & -31651 $\pm$ 155\\
    & & 
    0.5 & \textbf{-30096 $\pm$ 290} & -29877 $\pm$ 298 & -29810 $\pm$ 304 & -29802 $\pm$ 308\\
    & & 
    1.0 & \textbf{-26975 $\pm$ 755} & -26397 $\pm$ 660 & -26469 $\pm$ 678 & -26400 $\pm$ 639\\
    & & 
    1.5 & \textbf{-24616 $\pm$ 1226} & -23743 $\pm$ 986 & -23646 $\pm$ 1063 & -23718 $\pm$ 1013\\
    & & 
    2.0 & \textbf{-22628 $\pm$ 1596} & -21119 $\pm$ 1434 & -21421 $\pm$ 1340 & -21322 $\pm$ 1411\\
    \bottomrule
    \end{tabular}
    }
    \end{center}
\end{table}

\begin{figure}[tb]
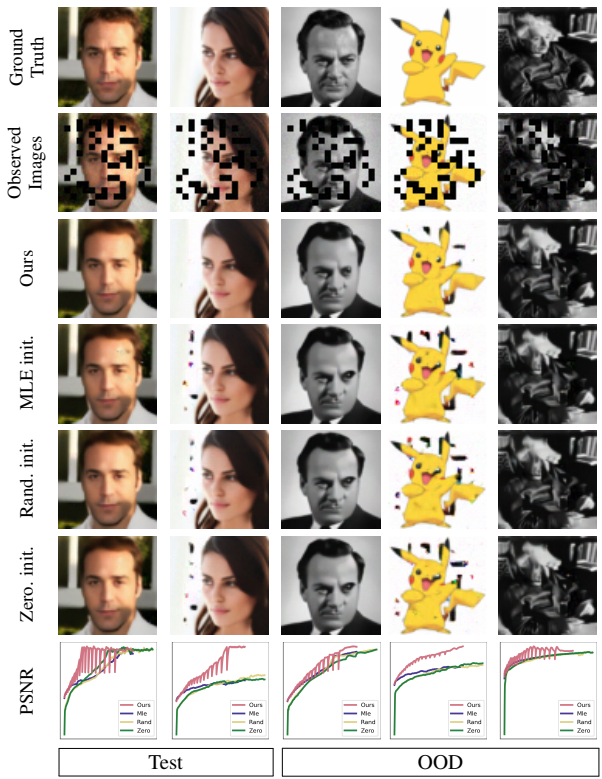

    \begin{center}

    \resizebox{\linewidth}{!}{
    \renewcommand{\tabcolsep}{2pt}
    \def\figwidth{0.2\linewidth}%
    \newcommand{\authornote}[1]{
        \footnotesize 
        \adjustbox{rotate=90}{\parbox{\figwidth}{\centering #1}}
    }
    \begin{tabular}{*{11}{c}}
      \authornote{Ground Truth}     &  
        \loadimage{figures/main/inpaint/glow}{width=\figwidth}{0}{5}\\
      \authornote{Observed Images}     & 
        \loadimage{figures/main/inpaint/glow}{width=\figwidth}{10}{5}\\
      \authornote{Ours}        & 
        \loadimage{figures/main/inpaint/glow}{width=\figwidth}{20}{5}\\
      \authornote{MLE init.}        & 
        \loadimage{figures/main/inpaint/glow}{width=\figwidth}{30}{5}\\
      \authornote{Rand. init.}          &  
        \loadimage{figures/main/inpaint/glow}{width=\figwidth}{40}{5}\\
      \authornote{Zero. init.}         &  
        \loadimage{figures/main/inpaint/glow}{width=\figwidth}{50}{5}\\
      \authornote{PSNR}         &  
        \loadimage{figures/main/inpaint/glow}{width=\figwidth}{60}{5}\\
      & \multicolumn{2}{c}{\fbox{\parbox{\calctotalwidth{\figwidth}{2}}{\centering Test}}} 
      & \multicolumn{3}{c}{\fbox{\parbox{\calctotalwidth{\figwidth}{3}}{\centering OOD}}}
    \end{tabular}
    }
    \caption{
    Results of inpainting CelebA faces and out-of-distribution images with GLOW as the generative prior. 
    Hyperparameter $\lambda=1.5$.
    }
    \label{fig:inpaint-glow}
    \end{center}
\end{figure}

We consider inpainting tasks that
assume $\yv = \Amat \xv + \ev$, where $\Amat \in \R^{n \times n}$ is a diagonal matrix with binary elements indicating whether a pixel is observed or missing.
In our experiments, we randomly mask several $4 \times 4$ pixel blocks such that in total the masked portion is 25\%. 
We sample $\ev \sim \gaus(0, 0.02^2\Imat)$ as the noise. 
Table \ref{tab:inpaint-loss} reports the MAP loss from different algorithms. Under all the settings, our {\notion} delivers much lower loss values. Again, the larger $\lambda$ is, the more apparent outperformance {\notion} exhibits. 

We show the true, observed, and reconstructed images with the GLOW prior in Figure \ref{fig:inpaint-glow}. The hyperparameter $\lambda$ is~$1.5$ such that all the algorithms achieve their best PSNRs. 
Specifically, for column~2, our algorithm removes all the masks whereas others fail to do so.
For column~4, {\notion} is the only one that recovers the main object of the image. These quantitative and visualization results together with those from the denoising and NCS tasks demonstrate the superiority of our amortized optimization in minimizing the MAP loss. 

In addition, checking the PSNR values in Tables~\ref{tab:realnvp-psnr} and~\ref{tab:glow-psnr} and the supplementary figures in the appendix, {\notion} in the denoising, NCS, and inpainting tasks using either RealNVP or GLOW results in outstanding reconstructions. On average, it gives higher PSNRs, 
recovers more details of the true images, and removes more noise.

%% file: subfiles/6_conclusion.tex
\section{Conclusions}
\label{sec:conclusion}
We propose {\notion}, a new amortized optimization algorithm for solving inverse problems with NFs as the generative prior, provide a theoretical guarantee, and  achieve remarkable improvement. 
The success of our method stems from two parts: 1) finding a good initialization and
2) easier and consecutive  optimizations that build up the original problem. 
The MLE initialization is indispensable for starting the sequence of subroutines. The amortization is the key contribution in that even with the same MLE  initialization, gradient descent without amortization yields a much worse performance than our method. We conclude that the {\notion} is an appealing solution to inverse problem optimization, considering its easy implementation, theoretical guarantee, and tremendous performance gains in various tasks. 
Our work implies that 
it is beneficial to design optimization algorithms that exploit the specific structure of the objective function before 
seeking gradient descent as a panacea. 

%% file: subfiles/7_appendix.tex
\clearpage

\title{Appendix}
\onecolumn

\section{Projected Gradient Descent for MLE}\label{app:projected}
We elaborate more details about solving the MLE $\xv^*(0)$ as defined in \eqref{eq:mle}. Note that in practice a valid image should always lie within the $[0, 1]^n$, and we include this consideration here. 

Let $\proj$ denote a projection operator and $\clip(x, a, b) = (x \land a) \lor b$ denote a standard clipping operation.

For denoising task when $f(\xv) = \xv$, the affine constraint contains only one single point $\xv = \yv$,
and the projection can be done by applying the following projection
\begin{align*}
    \proj\nolimits_{[0, 1]^n}(\xv) = (\clip(x_1, 0, 1),  \dots, \clip(x_n, 0, 1))^\top. 
\end{align*}

For NCS tasks, $f(\xv) = \Amat \xv$ where $\Amat$ has full row rank. We apply the following iterative projection  \citep{beck2017first} along two constraints by having $\xv_0 = \xv$ and repeating the following steps until converges
\begin{align*}
\xv_{i} &= \xv_{i-1} - \Amat^\top (\Amat\Amat^\top)^{-1}(\Amat \xv_{i-1} - \yv ) \\
\xv_{i} &=  \proj\nolimits_{[0, 1]^n}(\xv_i).
\end{align*}

For inpainting tasks, $f(\xv) = \Amat \xv$ where $\Amat$ is a diagonal matrix with 0/1 entries at its diagonal. Here 1 indicates that the pixel is observed and 0 indicates the opposite. Again we apply the following iterative projection \citep{beck2017first} along two constraints by having $\xv_0 = \xv$ and repeating the following steps until converges
\begin{align*}
\xv_i &= \Amat \yv + (\Imat - \Amat) \xv_{i-1} \\
\xv_{i} &=  \proj\nolimits_{[0, 1]^n}(\xv_i).
\end{align*}

\section{Omitted Proofs}

\subsection{Proof of Remark \ref{rmk:asmp_local}}
\label{sec_app:pf_rmk}
\begin{proof}[Proof of Remark \ref{rmk:asmp_local}]
(i) If $\exists \mu'>0$, $\exists \delta>0$, s.t.$\forall \lambda\in [0,\Lambda]$, $\mathcal{L}_{MAP}(\cdot,\lambda)$ is $\mu'$-strongly convex on $B(\xv^*(\lambda),\delta)$, then $\forall \xv \in B(\xv^*(\lambda),\delta)$, we have
\begin{align*}
    &\mathcal{L}_{MAP}(\xv,\lambda)\\
    \geq &\mathcal{L}_{MAP}(\xv^*,\lambda)\\
    \geq & \mathcal{L}_{MAP}(\xv,\lambda)+\nabla_\xv  \mathcal{L}_{MAP}(\xv,\lambda)^\intercal (\xv^*(\lambda)-\xv)+\frac{\mu'}{2}\norm{\xv^*(\lambda)-\xv}^2\\
    \geq&\mathcal{L}_{MAP}(\xv,\lambda)-\norm{\nabla_\xv
    \mathcal{L}_{MAP}(\xv,\lambda)}\norm{\xv^*(\lambda)-\xv}+\frac{\mu'}{2}\norm{\xv^*(\lambda)-\xv}^2.
\end{align*}

Therefore, we have
\begin{equation*}
    -\norm{\nabla_\xv
    \mathcal{L}_{MAP}(\xv,\lambda)}\norm{\xv^*(\lambda)-\xv}\\ +\frac{\mu'}{2}\norm{\xv^*(\lambda)-\xv}^2\leq 0,
\end{equation*}
from which we can see that
\begin{equation*}
    \norm{\xv^*(\lambda)-\xv}\leq \frac{2}{\mu'}\norm{\nabla_\xv
    \mathcal{L}_{MAP}(\xv,\lambda)}.
\end{equation*}

(ii) $\forall \xv \in B(\xv^*(\lambda),\delta)$, $\exists t\in [0,1]$, $\xiv=t\xv^*(\lambda)+(1-t)\xv$, s.t.
\begin{align*}
    &\mathcal{L}_{MAP}(\xv,\lambda)-\mathcal{L}_{MAP}(\xv^*,\lambda)\\
    =&\nabla_\xv \mathcal{L}_{MAP}(\xiv,\lambda)^\intercal (\xv-\xv^*(\lambda))\\
    \leq & \norm{\nabla_\xv \mathcal{L}_{MAP}(\xiv,\lambda)}\norm{\xv-\xv^*(\lambda)}\\
    =&\norm{\nabla_\xv \mathcal{L}_{MAP}(\xiv,\lambda)-\nabla_\xv \mathcal{L}_{MAP}(\xv^*,\lambda)}\norm{\xv-\xv^*(\lambda)}\\
    \leq & L\norm{\xiv-\xv^*(\lambda)}\norm{\xv-\xv^*(\lambda)}\\
    \leq & L\norm{\xv-\xv^*(\lambda)}^2\\
    \leq &L\sigma^2\norm{\nabla_\xv \mathcal{L}_{MAP}(\xv,\lambda)}^2.
\end{align*}
\end{proof}

\subsection{Proof of Theorem \ref{thm:cascade}}
\label{sec_app:pf_thm}
The proof of Theorem \ref{thm:cascade} relies on the following three lemmas:
\begin{lemma}\label{lm_app:limit}
    If Assumption~\ref{asmp:C_smooth_of_x} holds, then $\lim_{\lambda\rightarrow 0^+} \xv^*(\lambda)$ exists and $\xv^*(0)=\lim_{\lambda\rightarrow 0^+} \xv^*(\lambda)$, where $\xv^*(0)$ is defined in \eqref{eq:mle}.
\end{lemma}
\begin{proof}[Proof of Lemma \ref{lm_app:limit}]
We show the existence of $\lim_{\lambda\rightarrow 0^+} \xv^*(\lambda)$ by first proving that $\xv^*(\lambda)$ is bounded on $(0,\Lambda]$. Otherwise, for any given $\lambda_1\in(0,\Lambda]$, there exists $\lambda_2\in(0,\lambda_1)$, s.t. $\norm{\xv^*(\lambda_2)}>\norm{\xv^*(\lambda_1)}+C\lambda_1$. Then we have
$$\norm{\xv^*(\lambda_1)-\xv^*(\lambda_2)}\geq \norm{\xv^*(\lambda_2)}-\norm{\xv^*(\lambda_1)}>C\lambda_1\geq C|\lambda_1-\lambda_2|\,.$$
This contradicts to the Lipschitz continuity of $\xv^*(\lambda)$ on $(0,\Lambda]$ in Assumption~\ref{asmp:C_smooth_of_x}. Therefore $\xv^*(\lambda)$ is bounded on $(0,\Lambda]$. By Bolzano-Weierstrass Theorem, for any sequence $\{\lambda_n\}$ on $(0,\Lambda]$ such that $\lim_{n\rightarrow +\infty}\lambda_n=0$, $\{\xv^*(\lambda_n)\}$ has a convergent subsequence $\{\xv^*(\lambda_{n_k})\}$. 
Denote $\lim_{k \rightharpoonup +\infty} \xv^*(\lambda_{n_k}) = \bar\xv$, 
then $\forall \varepsilon>0$, $\exists k \in \N_+ $, s.t. 
$\lambda_{n_k}\in(0,\frac{\epsilon}{2C}]$ and $\norm{\xv^*(\lambda_{n_k})-\bar\xv}<\frac{\varepsilon}{2}$ hold. By the Lipschitz continuity of $\xv^*(\lambda)$, we have 
$\forall\lambda\in(0,\lambda_{n_k}]$
\begin{align*}
    \norm{\xv^*(\lambda)-\bar\xv}
    \leq \norm{\xv^*(\lambda)-\xv^*(\lambda_{n_k})}+\norm{\xv^*(\lambda_{n_k})-\bar\xv}
    \leq C|\lambda-\lambda_{n_k}|+\frac{\varepsilon}{2}
    \leq C\lambda_{n_k}+\frac{\varepsilon}{2}\leq\varepsilon\,.
\end{align*}
This indicates that $\lim_{\lambda\rightarrow 0^+} \xv^*(\lambda)=\bar\xv$.


If $f(\bar\xv)\ne\yv$, since $-\log p_e(\yv - f(\xv))$ reaches its minimum value if and only if $\yv = f(\xv)$, we have 
$$\triangle_1:=\log p_e(\yv - f(\xv^*(0)))-\log p_e(\yv - f(\bar\xv))>0.$$

By the continuity of $\log p_e(\yv - f(\xv))$ and $\log p_G(\xv)$ in $\xv$, there exists small enough $\lambda>0$, s.t.
$$\log p_e(\yv - f(\xv^*(\lambda)))<\log p_e(\yv - f(\bar\xv))+\frac{\triangle_1}{2}\,,$$
and
$$\lambda (\log p_G(\xv^*(\lambda))-\log p_G(\xv^*(0)))<\frac{\triangle_1}{2}\,.$$
Combining the above inequalities, we have
\begin{align*}
    &\log p_e(\yv - f(\xv^*(\lambda)))+\lambda \log p_G(\xv^*(\lambda))\\
    <&\log p_e(\yv - f(\bar\xv))+\lambda \log p_G(\xv^*(0))+\triangle_1\\
    =&\log p_e(\yv - f(\xv^*(0)))+\lambda \log p_G(\xv^*(0))\,.
\end{align*}
This contradicts to the definition of $\xv^*(\lambda)$ (see Eq.~\eqref{eq:map}). This contradiction indicates $f(\bar\xv)$ must be equal to $\yv$, which implies that $\bar\xv$ is a feasible solution of problem~\eqref{eq:mle}. So
if $\bar\xv$ is not an optimal solution of problem~\eqref{eq:mle}, then we have
$$\log p_G(\xv^*(0))>\log p_G(\bar\xv)\,.$$
Thus by the continuity of $\log p_G$, there exists small enough $\lambda>0$, s.t.
\begin{equation}\label{eq_app:ineq1}
    \log p_G(\xv^*(\lambda))<\log p_G(\xv^*(0))\,.
\end{equation}

Since $f(\xv^*(0)))=\yv$, we have
\begin{equation}\label{eq_app:ineq2}
    \log p_e(\yv - f(\xv^*(0)))\geq\log p_e(\yv - f(\xv^*(\lambda)))\,.
\end{equation}

Combining \eqref{eq_app:ineq1} and \eqref{eq_app:ineq2}, we obtain
\begin{align*}
    \log p_e(\yv - f(\xv^*(\lambda)))+\lambda \log p_G(\xv^*(\lambda))
    <\log p_e(\yv - f(\xv^*(0)))+\lambda\log p_G(\xv^*(0))\,,
\end{align*}
which again contradicts to the definition of $\xv^*(\lambda)$.

Thus $\bar\xv=\xv^*(0)$.
\end{proof}

Lemma~\ref{lm_app:limit} indicates that when Assumption~\ref{asmp:C_smooth_of_x} holds, $\xv^*(\lambda)$ is Lipschitz continuous on $[0,\Lambda]$, i.e., for all $\lambda_1,\lambda_2\in [0,\Lambda]$, $\norm{\xv^*(\lambda_1)- \xv^*(\lambda_2)}\leq C|\lambda_1-\lambda_2|.$

\begin{lemma}\label{lm_app:F_decrease}
    If Assumption \ref{asmp:L_smooth_of_F} holds, then $\forall \lambda\in [0,\Lambda]$, let
    \begin{equation}\label{eq_app:x}
        \xv_{k+1}=\xv_k-\frac{1}{L}\nabla_\xv \mathcal{L}_{MAP}(\xv_k,\lambda).
    \end{equation}
    
    We have
    \begin{equation}\label{eq_app:1}
        \mathcal{L}_{MAP}(\xv_{k+1},\lambda)\leq \mathcal{L}_{MAP}(\xv_k,\lambda)-\frac{1}{2L}\norm{\nabla_\xv \mathcal{L}_{MAP}(\xv_k,\lambda)}^2\,.
    \end{equation}
\end{lemma}

\begin{proof}[Proof of Lemma \ref{lm_app:F_decrease}]
\begin{align*}
    &\mathcal{L}_{MAP}(\xv_{k+1},\lambda)\\
    \leq &\mathcal{L}_{MAP}(\xv_k,\lambda)+\nabla_\xv \mathcal{L}_{MAP}(\xv_k,\lambda)^\intercal (\xv_{k+1}-\xv_k)+\frac{L}{2}\norm{\xv_{k+1}-\xv_k}^2\\
    =&\mathcal{L}_{MAP}(\xv_k,\lambda)-\frac{1}{L}\norm{\nabla_\xv \mathcal{L}_{MAP}(\xv_k,\lambda)}^2+\frac{1}{2L}\norm{\nabla_\xv \mathcal{L}_{MAP}(\xv_k,\lambda)}^2\\
    =&\mathcal{L}_{MAP}(\xv_k,\lambda)-\frac{1}{2L}\norm{\nabla_\xv \mathcal{L}_{MAP}(\xv_k,\lambda)}^2\,.
\end{align*}
\end{proof}

\begin{lemma}\label{lm_app:local}
    Assume Assumption \ref{asmp:L_smooth_of_F} and \ref{asmp:local} hold. Let $\mu=\frac{1}{2L\sigma^2}$. For any fixed $\delta >0$, let
    \begin{equation*}
        \delta_0=\min\left\{\delta,\frac{\mu}{\sqrt{2(L-\mu)L}}\delta\right\}.
    \end{equation*}
    If $\xv_0\in B(\xv^*(\lambda),\delta_0)$, and $\{\xv_k\}$ is generated by \eqref{eq_app:x}, then $\forall k\in\N_+$, $\xv_k\in B(\xv^*(\lambda),\delta)$.
\end{lemma}

\begin{proof}[Proof of Lemma \ref{lm_app:local}]
We prove by induction on $k$. when $k=0$, we have
\begin{equation*}
    \xv_0\in B(\xv^*(\lambda),\delta_0)\subset B(\xv^*(\lambda),\delta).
\end{equation*}

Assume when $k\geq 1$, $\xv_j\in B(\xv^*(\lambda),\delta)$, $\forall j\leq k$, then by \eqref{eq_app:1} we have
\begin{align*}
    &\mathcal{L}_{MAP}(\xv_k,\lambda)-\mathcal{L}_{MAP}(\xv^*(\lambda),\lambda)\\
    \leq &\mathcal{L}_{MAP}(\xv_{k-1},\lambda)-\mathcal{L}_{MAP}(\xv^*(\lambda),\lambda)-\frac{1}{2L}\norm{\nabla_\xv \mathcal{L}_{MAP}(\xv_{k-1},\lambda)}^2.
\end{align*}

By induction hypothesis and \eqref{eq:2}, we have
\begin{align*}
    &\mathcal{L}_{MAP}(\xv_{k-1},\lambda)-\mathcal{L}_{MAP}(\xv^*(\lambda),\lambda)-\frac{1}{2L}\norm{\nabla_\xv \mathcal{L}_{MAP}(\xv_{k-1},\lambda)}^2\\
    \leq &(1-\frac{\mu}{L})(\mathcal{L}_{MAP}(\xv_{k-1},\lambda)-\mathcal{L}_{MAP}(\xv^*(\lambda),\lambda)).
\end{align*}

From the above two inequalities we deduce
\begin{equation}\label{eq_app:recursion}
\begin{split}
    &\mathcal{L}_{MAP}(\xv_{k},\lambda)-\mathcal{L}_{MAP}(\xv^*(\lambda),\lambda)\\
    \leq &(1-\frac{\mu}{L})(\mathcal{L}_{MAP}(\xv_{k-1},\lambda)-\mathcal{L}_{MAP}(\xv^*(\lambda),\lambda))\\
    \leq &\cdots\\
    \leq &(1-\frac{\mu}{L})^k(\mathcal{L}_{MAP}(\xv_0,\lambda))-\mathcal{L}_{MAP}(\xv^*(\lambda),\lambda)).
\end{split}
\end{equation}
From \eqref{eq_app:1} we can see that
\begin{equation}\label{eq_app:3}
     \mathcal{L}_{MAP}(\xv^*(\lambda),\lambda)\leq \mathcal{L}_{MAP}(\xv_k,\lambda)-\frac{1}{2L}\norm{\nabla_\xv \mathcal{L}_{MAP}(\xv_k,\lambda)}^2.
\end{equation}

Thus we have
\begin{align*}
    &\mu \norm{\xv_k-\xv^* (\lambda)}^2\\
    \overset{(a)}{\leq} &\frac{1}{2L}\norm{\nabla_\xv \mathcal{L}_{MAP}(\xv_k,\lambda)}^2\\
    \overset{(b)}{\leq}&\mathcal{L}_{MAP}(\xv_k,\lambda)-\mathcal{L}_{MAP}(\xv^*(\lambda),\lambda)\\
    \overset{(c)}{\leq}&(1-\frac{\mu}{L})^k(\mathcal{L}_{MAP}(\xv_0,\lambda)-\mathcal{L}_{MAP}(\xv^*(\lambda),\lambda))\\
    \overset{(d)}{\leq}&(1-\frac{\mu}{L})^k\frac{1}{2\mu}\norm{\nabla_\xv \mathcal{L}_{MAP}(\xv_0,\lambda)}^2\\
    \overset{(e)}{\leq}&(1-\frac{\mu}{L})^k\frac{L^2}{2\mu}\norm{\xv_0-\xv^*(\lambda)}^2,
\end{align*}
where step (a) is by Assumption \ref{asmp:local} (recall that $\mu=\frac{1}{2L\sigma^2}$), step (b), step (c), and step (d) are based on equation \eqref{eq_app:3}, equation \eqref{eq_app:recursion}, and equation \eqref{eq:2} respectively. Step (e) is by Assumption \ref{asmp:L_smooth_of_F}.

From the above inequalities, we deduce
\begin{equation}\label{eq_app:star}
    \norm{\xv_k-\xv^*(\lambda)}
    \leq\frac{L}{\sqrt{2}\mu}(1-\frac{\mu}{L})^{k/2}\norm{\xv_0-\xv^*(\lambda)}
    \leq\frac{\delta}{2}.
\end{equation}
Note that
\begin{align*}
    \norm{\xv_{k+1}-\xv_k}
    &=\frac{1}{L}\norm{\nabla_\xv \mathcal{L}_{MAP}(\xv_k,\lambda)}\\
    &=\frac{1}{L}\norm{\nabla_\xv \mathcal{L}_{MAP}(\xv_k,\lambda)-\nabla_\xv \mathcal{L}_{MAP}(\xv^*,\lambda)}\\
    &\leq\norm{\xv_k-\xv^*(\lambda)}\leq\frac{\delta}{2}.
\end{align*}

So we have 
\begin{equation*}
    \norm{\xv_{k+1}-\xv^*(\lambda)}
    \leq \norm{\xv_k-\xv^*(\lambda)}+    \norm{\xv_{k+1}-\xv_k}
    \leq \delta.
\end{equation*}

The induction step is complete.
\end{proof}

Now we are ready to prove the Theorem \ref{thm:cascade}.
\begin{proof}[Proof of Theorem \ref{thm:cascade}]
From the first inequality of \eqref{eq_app:star} we know that if $\xv_0\in B(\xv^*(\lambda),\delta_0)$, then
\begin{equation}\label{eq_app:tighter}
    \norm{\xv_k-\xv^*(\lambda)}\leq(1-\frac{\mu}{L})^{k/2}\delta,\,\forall k\in \mathbb{N}.
\end{equation}

Let $\lambda_0=0$, $\lambda_{i+1}=\lambda_{i}+\frac{\Lambda}{N}$, $i=1,2,\cdots,N-1$, where $N\geq\frac{2C\Lambda}{\delta_0}$. Let $\varDelta \lambda\triangleq\frac{\Lambda}{N}$. Suppose $\xv_k(\lambda_i)$ is the point in the last iteration of the inner loop when $\lambda=\lambda_i$, where $K=[\frac{2\log(2\delta/\delta_0)}{\log(L/(L-\mu))}]+1.$ 

Let $\xv_0(\lambda_0)=G^{-1}(\yv)$, then we have
\begin{equation}\label{eq_app:0start}
    \norm{\xv_0(\lambda_{0})-\xv^*(\lambda_{0})}=0.
\end{equation}

If $\norm{\xv_0(\lambda_i)-\xv^*(\lambda_i)}\leq\delta_0$, then by \eqref{eq_app:tighter} we deduce
\begin{equation*}
    \norm{\xv_K(\lambda_i)-\xv^*(\lambda_i)}\leq\delta.
\end{equation*}

When $\lambda=\lambda_{i+1}$, let $\xv_0(\lambda_{i+1})=\xv_K(\lambda_i)$, then we have
\begin{align*}
    &\norm{\xv_0(\lambda_{i+1})-\xv^*(\lambda_{i+1})}\\
    \leq&\norm{\xv_0(\lambda_{i+1})-\xv^*(\lambda_{i})}+\norm{\xv^*(\lambda_{i})-\xv^*(\lambda_{i+1})}\\
    \overset{(a)}{\leq}&\norm{\xv_K(\lambda_i)-\xv^*(\lambda_i)}+C\varDelta\lambda\\
    \leq&\delta_0,
\end{align*}
where (a) uses Assumption \ref{asmp:C_smooth_of_x} and $\xv_0(\lambda_{i+1})=\xv_K(\lambda_i)$.

Thus by \eqref{eq_app:0start} and induction we deduce
\begin{equation*}
    \norm{\xv_0(\lambda_{i})-\xv^*(\lambda_{i})}\leq \delta_0,\,\forall i\in\{0,1,\cdots,N\}.
\end{equation*}

When $i=N$, $K=\max\{0,[\frac{2\log(2\varepsilon/\delta_0)}{\log(L/(L-\mu))}]+1\}$, again using the first inequality in \eqref{eq_app:star}, we have
\begin{equation*}
    \norm{\xv_K(\Lambda)-\xv^*(\Lambda)}\leq \varepsilon.
\end{equation*}
\end{proof}

\section{Alternative Amortized Optimization for Inverse Problems}
In this section we present an alternative version of Amortized Optimization for $\xv$, which is used in experiments.
In practice we adopt the following hyper-parameters as summarized in Table \ref{tab:hparams}.

\begin{remark}
In practice it is {difficult} to quantify the \textit{gradual change},
so in step 8, we determine the step size $\delta$ such that the relative change of the loss is not too large,
and for practical concern, we also want the step size not too small. 
In step 9 and 10, 
we update our belief of whether the minimal step size is too aggressive.
We check how much the MAP loss decreases in this round compared with the previous one. 
Intuitively, if current change is larger, then $\lambda$ has been updated too much such that the previous solution, $\xv_0$ in this round, is far from $\xv^*(\lambda)$, then we decrease the $\delta_{\min}$ by raising it to power $\Delta^\prime_{\loss} / \Delta_{\loss}$, the prespecified bound will guarantee the result decrease, and vice versa. 
Finally, we will end up with a minimal step size $\delta_{\min}$ that allows us to achieve a stable loss change rate.
We found this algorithm works empirically well.  
\end{remark}

\begin{algorithm}[H]
    \caption{Alternative {\notion} algorithm}
    \label{alg:cascadex2}
    \begin{algorithmic}[1] 
        \STATE \textbf{Input:} observation $\yv$, generative model $G$, targeted hyperparameter $\Lambda > 0$, target change rate of loss magnitude $r \in (0, 1)$, minimum step size $\delta_{\min}^0$ and its bound $1 \geq \delta_{\min, h} \geq \delta_{\min, l} > 0$ for updating $\lambda$, maximum iteration number $K > 0$, step size $\alpha > 0$ for solving $\mathcal{L}_\text{MAP}(\xv, \lambda)$. 
        \STATE \textbf{Initialize:}
        $\lambda=0, 
        \xv_0 = f^{-1}(\yv), 
        \delta_{\min} = \delta_{\min}^0,
        \Delta_{\loss} = \texttt{NULL}$
        \REPEAT  
        \FOR{$k=1, \dots, K$} 
        \STATE $\xv_k=\xv_{k-1} - \alpha \nabla_{\xv}\mathcal{L}_\text{MAP}(\xv_{k-1}, \lambda)$
        \ENDFOR
        \STATE
        $\ev = \yv - f(\xv_{K})$
        \STATE
        Determine the step size $\delta$ 
        such that 
        \begin{align*}
            r = \left|
            \frac{\mathcal{L}_\text{MAP}(\xv_{K}; \lambda + \delta) - \mathcal{L}_\text{MAP}(\xv_{K}; \lambda)}
            {\mathcal{L}_\text{MAP}(\xv_{K}; \lambda) } 
            \right| 
            \  
            \Rightarrow 
            \  
            \delta =  r \left|
            \frac{\log p_e(\ev)}{\log p_G(\xv_{K})} + \lambda
            \right|
        \end{align*}
        while satisfying $\delta \geq \delta_{\min}$,
        i.e., 
        \begin{align*}
            \delta
            = 
            \delta \land \delta_{\min}
        \end{align*}
        \IF {$\Delta_{\loss} \neq \texttt{NULL}$}
            \STATE
            Update $\delta_{\min}$ based on the relative loss change  
            \begin{align*}
            \Delta^\prime_{\loss} = \left| 
            \frac{\loss_{\text{MAP}}(\xv_0, \lambda) - \loss_{\text{MAP}}(\xv_K, \lambda)} {\loss_{\text{MAP}}(\xv_0, \lambda)}
            \right|
            \end{align*}
            within the specified range
            \begin{align*}
            \delta_{\min} = \clip( \Delta^\prime_{\loss} / \Delta_{\loss}, \delta_{\min, l},  \delta_{\min, h} ) 
            \end{align*}
        \ENDIF
        \STATE
        Update 
        $\Delta_{\loss} = \Delta^\prime_{\loss}$, $K = 1.1 K$, $\alpha = 0.99 \alpha$.
        \STATE
        $\lambda = \lambda + \delta$
        \STATE
        $\xv_0 = \xv_{K}$
        \UNTIL{$\lambda \geq \Lambda$}
        \OUTPUT $\xv_{K}$
    \end{algorithmic}
\end{algorithm}

\begin{table*}[!tb]
    \begin{center}
    \caption{
    Hyper-parameters used in Amortized Optimization appeared Algorithm \ref{alg:cascadex2}.
    }
    \label{tab:hparams}
    \begin{tabular}{rcccccc}
    \toprule
     Hyperparameter & step size $\alpha$ & iteration $K$ & target rate $r$ & min. step size $\delta^0_{\min}$ & $\delta_{\min, h}$ & $\delta_{\min, h}$ \\
     \midrule
     Value &  0.05 & 40 & 0.05 & $\Lambda / 20$ &  $4\delta^0_{\min} \lor 1$ & $\delta^0_{\min} / 4$ \\
    \bottomrule
    \end{tabular}
    \end{center}
\end{table*}

\section{Additional Experiment Results}

In this section we report more experiment results, including PSNR values on all tasks, and more reconstructed images using the hyperparameter $\lambda$ based on which most algorithms can achieve the highest PSNR.

\begin{table*}[!tb]
    \scriptsize   
    \begin{center}
    \caption{
     PSNR (mean$\pm$se) of different algorithms using RealNVP as the generative prior, higher is better.
    Best results are in {bold}.
    }
    \label{tab:realnvp-psnr}
    \resizebox{0.8\linewidth}{!}{
    \begin{tabular}{rrccccc}
    \toprule
     & Algorithm & $\lambda=0.3$ & $\lambda=0.5$ & $\lambda=1.0$  & $\lambda=1.5$ & $\lambda=2.0$ \\
     
    \midrule
    \multicolumn{7}{c}{Denoising} \\
    \midrule
    \multirow{4}{*}{\rotatebox[origin=c]{90}{Test}}
    & {Ours} &  
    \textbf{29.92 $\pm$ 0.08} & \textbf{30.02 $\pm$ 0.06} & \textbf{28.27 $\pm$ 0.06} & \textbf{27.04 $\pm$ 0.06} & \textbf{26.11 $\pm$ 0.06}\\\
    & MLE. init. & 
    29.74 $\pm$ 0.08 & 29.75 $\pm$ 0.07 & 28.13 $\pm$ 0.07 & 27.00 $\pm$ 0.07 & 26.09 $\pm$ 0.07\\
    & Rand. init. & 
    29.62 $\pm$ 0.08 & 29.72 $\pm$ 0.07 & 28.14 $\pm$ 0.07 & 26.91 $\pm$ 0.08 & \textbf{26.11 $\pm$ 0.07}\\
    & Zero. init. & 
    29.60 $\pm$ 0.08 & 29.71 $\pm$ 0.07 & 28.14 $\pm$ 0.07 & 26.96 $\pm$ 0.07 & 26.04 $\pm$ 0.07\\
    \cmidrule{2-7}
    \multirow{4}{*}{\rotatebox[origin=c]{90}{OOD}}
    & Ours & 
    \textbf{28.70 $\pm$ 0.64} & \textbf{29.05 $\pm$ 0.55} & 27.25 $\pm$ 0.50 & 26.09 $\pm$ 0.54 & 25.36 $\pm$ 0.53\\
    & MLE. init. & 
    28.64 $\pm$ 0.67 & 28.95 $\pm$ 0.71 & \textbf{27.44 $\pm$ 0.69} & 26.25 $\pm$ 0.65 & \textbf{25.45 $\pm$ 0.60}\\
    & Rand. init. & 
    28.57 $\pm$ 0.68 & 28.83 $\pm$ 0.67 & 27.33 $\pm$ 0.65 & \textbf{26.29 $\pm$ 0.67} & 25.44 $\pm$ 0.61\\
    & Zero. init. & 
    28.53 $\pm$ 0.65 & 28.89 $\pm$ 0.72 & 27.42 $\pm$ 0.68 & \textbf{26.29 $\pm$ 0.63} & 25.38 $\pm$ 0.57\\

    \midrule
    \multicolumn{7}{c}{{Noisy Compressed Sensing} when $m=4000$} \\
    \midrule
    \multirow{4}{*}{\rotatebox[origin=c]{90}{{Test}}}
    & Ours & 
    \textbf{29.53 $\pm$ 0.09} & \textbf{29.28 $\pm$ 0.07} & \textbf{27.70 $\pm$ 0.07} & \textbf{26.60 $\pm$ 0.07} & 25.69 $\pm$ 0.07\\
    & MLE. init. & 
    29.14 $\pm$ 0.10 & 28.93 $\pm$ 0.08 & 27.54 $\pm$ 0.08 & 26.43 $\pm$ 0.08 & 25.67 $\pm$ 0.08\\
    & Rand. init. & 
    29.37 $\pm$ 0.08 & 29.03 $\pm$ 0.08 & 27.60 $\pm$ 0.08 & 26.52 $\pm$ 0.08 & 25.69 $\pm$ 0.07\\
    & Zero. init. & 
    29.39 $\pm$ 0.08 & 29.01 $\pm$ 0.08 & 27.60 $\pm$ 0.08 & 26.47 $\pm$ 0.07 & \textbf{25.74 $\pm$ 0.08}\\
    \cmidrule{2-7}
    \multirow{4}{*}{\rotatebox[origin=c]{90}{OOD}}
    & Ours & 
    28.11 $\pm$ 0.83 & \textbf{28.19 $\pm$ 0.68} & 26.72 $\pm$ 0.68 & \textbf{25.82 $\pm$ 0.70} & \textbf{24.92 $\pm$ 0.54}\\
    & MLE. init. & 
    27.77 $\pm$ 0.94 & 27.80 $\pm$ 0.84 & 26.58 $\pm$ 0.87 & 25.32 $\pm$ 0.84 & 24.72 $\pm$ 0.68\\
    & Rand. init. & 
    \textbf{28.19 $\pm$ 0.79} & 28.09 $\pm$ 0.73 & \textbf{26.81 $\pm$ 0.78} & 25.56 $\pm$ 0.73 & 24.80 $\pm$ 0.67\\
    & Zero. init. & 
    27.99 $\pm$ 0.88 & 28.10 $\pm$ 0.80 & 26.93 $\pm$ 0.73 & 25.61 $\pm$ 0.69 & 24.87 $\pm$ 0.66\\

    \midrule
    \multicolumn{7}{c}{{Noisy Compressed Sensing} when $m=2000$} \\
    \midrule
    \multirow{4}{*}{\rotatebox[origin=c]{90}{Test}}
    & Ours & 
    \textbf{28.90 $\pm$ 0.09} & \textbf{28.48 $\pm$ 0.08} & \textbf{27.17 $\pm$ 0.07} & \textbf{26.15 $\pm$ 0.07} & \textbf{25.42 $\pm$ 0.07}\\
    & MLE. init. & 
    28.52 $\pm$ 0.10 & 28.20 $\pm$ 0.09 & 27.03 $\pm$ 0.09 & 26.08 $\pm$ 0.08 & 25.37 $\pm$ 0.08\\
    & Rand. init. & 
    28.73 $\pm$ 0.09 & 28.30 $\pm$ 0.08 & 27.04 $\pm$ 0.08 & 26.08 $\pm$ 0.08 & 25.31 $\pm$ 0.08\\
    & Zero. init. & 
    28.72 $\pm$ 0.09 & 28.28 $\pm$ 0.09 & 27.06 $\pm$ 0.08 & {26.11 $\pm$ 0.08} & 25.30 $\pm$ 0.07\\
    \cmidrule{2-7}
    \multirow{4}{*}{\rotatebox[origin=c]{90}{OOD}}
    & Ours & 
    27.21 $\pm$ 0.92 & \textbf{27.26 $\pm$ 0.81} & 25.92 $\pm$ 0.74 & \textbf{25.30 $\pm$ 0.62} &\textbf{ 24.57 $\pm$ 0.66}\\
    & MLE. init. & 
    26.85 $\pm$ 1.01 & 26.63 $\pm$ 1.02 & 25.79 $\pm$ 0.92 & 25.00 $\pm$ 0.82 & 24.10 $\pm$ 0.86\\
    & Rand. init. & 
    \textbf{27.29 $\pm$ 0.85} & 27.10 $\pm$ 0.84 & 25.99 $\pm$ 0.81 & 25.08 $\pm$ 0.78 & 24.37 $\pm$ 0.73\\
    & Zero. init. & 
    27.27 $\pm$ 0.90 & 27.09 $\pm$ 0.95 & \textbf{26.20 $\pm$ 0.79} & 25.24 $\pm$ 0.75 & 24.45 $\pm$ 0.68\\

    \midrule
    \multicolumn{7}{c}{Inpainting} \\
    \midrule
    \multirow{4}{*}{\rotatebox[origin=c]{90}{Test}}
    & Ours & 
    \textbf{32.73 $\pm$ 0.14} &\textbf{ 33.06 $\pm$ 0.15} & \textbf{33.19 $\pm$ 0.14} & \textbf{32.83 $\pm$ 0.14} & \textbf{32.50 $\pm$ 0.13}\\
    & MLE. init. & 
    31.80 $\pm$ 0.18 & 32.10 $\pm$ 0.18 & 32.02 $\pm$ 0.19 & 31.69 $\pm$ 0.18 & 31.52 $\pm$ 0.17\\
    & Rand. init. & 
    31.30 $\pm$ 0.21 & 31.34 $\pm$ 0.22 & 31.33 $\pm$ 0.22 & 31.38 $\pm$ 0.20 & 31.31 $\pm$ 0.18\\
    & Zero. init. & 
    31.21 $\pm$ 0.22 & 31.33 $\pm$ 0.23 & 31.42 $\pm$ 0.21 & 31.55 $\pm$ 0.20 & 31.20 $\pm$ 0.18\\
    \cmidrule{2-7}
    \multirow{4}{*}{\rotatebox[origin=c]{90}{OOD}}
    & Ours & 
    \textbf{28.68 $\pm$ 1.84} & \textbf{28.99 $\pm$ 1.84} & \textbf{28.86 $\pm$ 1.80} & \textbf{29.13 $\pm$ 1.62} &\textbf{ 28.57 $\pm$ 1.42}\\
    & MLE. init. & 
    27.81 $\pm$ 2.05 & 28.02 $\pm$ 1.94 & 27.83 $\pm$ 1.96 & 27.75 $\pm$ 2.03 & 28.20 $\pm$ 1.75\\
    & Rand. init. & 
    27.63 $\pm$ 2.06 & 28.03 $\pm$ 2.20 & 27.96 $\pm$ 2.20 & 27.77 $\pm$ 1.93 & 27.83 $\pm$ 1.98\\
    & Zero. init. & 
    27.55 $\pm$ 2.11 & 27.98 $\pm$ 2.12 & 27.92 $\pm$ 2.09 & 27.95 $\pm$ 1.96 & 27.93 $\pm$ 1.98\\    
    \bottomrule
    \end{tabular}
    }
    \end{center}
\end{table*}

\begin{table*}[!tb]
    \scriptsize   
    \begin{center}
    \caption{
    PSNR (mean$\pm$se) of different algorithms using GLOW as the generative prior, higher is better.
    Best results are in {bold}.
    }
    \label{tab:glow-psnr}
    \resizebox{0.8\linewidth}{!}{
    \begin{tabular}{rrccccc}
    \toprule
     & Algorithm & $\lambda=0.3$ & $\lambda=0.5$ & $\lambda=1.0$  & $\lambda=1.5$ & $\lambda=2.0$ \\
     
    \midrule
    \multicolumn{7}{c}{{Denoising}} \\
    \midrule
    \multirow{4}{*}{\rotatebox[origin=c]{90}{Test}}
    & {Ours} & 
    29.32 $\pm$ 0.09 & 29.29 $\pm$ 0.06 & 27.29 $\pm$ 0.07 & 25.88 $\pm$ 0.07 & 24.81 $\pm$ 0.07\\
    & MLE. init. & 
    29.46 $\pm$ 0.09 & 29.37 $\pm$ 0.06 & 27.41 $\pm$ 0.07 & \textbf{25.96 $\pm$ 0.07} & \textbf{24.90 $\pm$ 0.07}\\
    & Rand. init. & 
    \textbf{29.51 $\pm$ 0.09} & \textbf{29.41 $\pm$ 0.06} & \textbf{27.42 $\pm$ 0.07} & 25.94 $\pm$ 0.07 & 24.85 $\pm$ 0.06\\
    & Zero. init. & 
    29.49 $\pm$ 0.09 & 29.38 $\pm$ 0.06 & 27.41 $\pm$ 0.06 & 25.94 $\pm$ 0.07 & 24.84 $\pm$ 0.07\\
    \cmidrule{2-7}
    \multirow{4}{*}{\rotatebox[origin=c]{90}{OOD}}
    & Ours & 
    28.72 $\pm$ 0.62 & 28.54 $\pm$ 0.63 & 26.36 $\pm$ 0.57 & 24.97 $\pm$ 0.48 & \textbf{24.05 $\pm$ 0.53}\\
    & MLE. init. & 
    28.70 $\pm$ 0.68 & 28.50 $\pm$ 0.56 & 26.29 $\pm$ 0.50 & \textbf{25.11 $\pm$ 0.50} & 23.98 $\pm$ 0.46\\
    & Rand. init. & 
    \textbf{28.94 $\pm$ 0.63} & 28.44 $\pm$ 0.56 & \textbf{26.52 $\pm$ 0.56} & 24.96 $\pm$ 0.47 & 23.94 $\pm$ 0.45\\
    & Zero. init. & 
    28.86 $\pm$ 0.64 & \textbf{28.56 $\pm$ 0.57} & 26.37 $\pm$ 0.55 & 25.01 $\pm$ 0.50 & 23.87 $\pm$ 0.49\\

    \midrule
    \multicolumn{7}{c}{{Noisy Compressed Sensing} when $m=4000$} \\
    \midrule
    \multirow{4}{*}{\rotatebox[origin=c]{90}{Test}}
    & Ours & 
    \textbf{28.78 $\pm$ 0.10} & \textbf{28.50 $\pm$ 0.07} & 26.80 $\pm$ 0.07 & 25.53 $\pm$ 0.07 & \textbf{24.51 $\pm$ 0.07}\\
    & MLE. init. & 
    28.38 $\pm$ 0.11 & 28.47 $\pm$ 0.08 & \textbf{26.82 $\pm$ 0.07} & \textbf{25.54 $\pm$ 0.08} & 24.49 $\pm$ 0.07\\
    & Rand. init. & 
    28.19 $\pm$ 0.11 & 28.43 $\pm$ 0.08 & 26.78 $\pm$ 0.07 & 25.37 $\pm$ 0.07 & 24.36 $\pm$ 0.07\\
    & Zero. init. & 
    28.20 $\pm$ 0.11 & 28.39 $\pm$ 0.07 & 26.74 $\pm$ 0.07 & 25.38 $\pm$ 0.07 & 24.31 $\pm$ 0.06\\
    \cmidrule{2-7}
    \multirow{4}{*}{\rotatebox[origin=c]{90}{OOD}}
    & Ours & 
    \textbf{27.78 $\pm$ 0.78} & \textbf{27.48 $\pm$ 0.72} & \textbf{25.80 $\pm$ 0.60} & \textbf{24.71 $\pm$ 0.54} & \textbf{23.70 $\pm$ 0.45}\\
    & MLE. init. & 
    27.05 $\pm$ 0.98 & 27.31 $\pm$ 0.79 & 25.65 $\pm$ 0.68 & 24.38 $\pm$ 0.57 & 23.48 $\pm$ 0.54\\
    & Rand. init. & 
    26.88 $\pm$ 0.95 & 27.29 $\pm$ 0.77 & 25.75 $\pm$ 0.64 & 24.38 $\pm$ 0.49 & 23.49 $\pm$ 0.48\\
    & Zero. init. & 
    26.84 $\pm$ 0.96 & 27.12 $\pm$ 0.76 & 25.68 $\pm$ 0.64 & 24.38 $\pm$ 0.48 & 23.53 $\pm$ 0.48\\
    
    \midrule
    
    \multicolumn{7}{c}{{Noisy Compressed Sensing} $m=2000$} \\
    \midrule
    \multirow{4}{*}{\rotatebox[origin=c]{90}{Test}}
    & Ours & 
    \textbf{28.02 $\pm$ 0.11} & \textbf{27.66 $\pm$ 0.09} & \textbf{26.26 $\pm$ 0.07} & \textbf{25.12 $\pm$ 0.07} & \textbf{24.15 $\pm$ 0.06}\\
    & MLE. init. & 
    25.04 $\pm$ 0.21 & 26.09 $\pm$ 0.16 & 25.68 $\pm$ 0.10 & 24.68 $\pm$ 0.08 & 23.75 $\pm$ 0.07\\
    & Rand. init. & 
    26.04 $\pm$ 0.16 & 26.75 $\pm$ 0.13 & 25.85 $\pm$ 0.08 & 24.65 $\pm$ 0.07 & 23.72 $\pm$ 0.07\\
    & Zero. init. & 
    25.76 $\pm$ 0.18 & 26.62 $\pm$ 0.14 & 25.82 $\pm$ 0.08 & 24.59 $\pm$ 0.07 & 23.71 $\pm$ 0.07\\
    \cmidrule{2-7}
    \multirow{4}{*}{\rotatebox[origin=c]{90}{OOD}}
    & Ours & 
    \textbf{26.64 $\pm$ 0.94} & \textbf{26.57 $\pm$ 0.78} & \textbf{25.33 $\pm$ 0.61} & \textbf{24.24 $\pm$ 0.57} & \textbf{23.42 $\pm$ 0.48}\\
    & MLE. init. & 
    22.19 $\pm$ 1.24 & 24.72 $\pm$ 1.30 & 24.12 $\pm$ 0.97 & 23.51 $\pm$ 0.80 & 22.82 $\pm$ 0.53\\
    & Rand. init. & 
    23.96 $\pm$ 1.44 & 25.26 $\pm$ 0.89 & 23.20 $\pm$ 1.38 & 23.87 $\pm$ 0.57 & 22.98 $\pm$ 0.51\\
    & Zero. init. & 
    23.71 $\pm$ 1.39 & 25.06 $\pm$ 1.13 & 23.42 $\pm$ 1.27 & 23.78 $\pm$ 0.60 & 22.86 $\pm$ 0.58\\
    
    \midrule
    \multicolumn{7}{c}{Inpainting} \\
    \midrule
    \multirow{4}{*}{\rotatebox[origin=c]{90}{Test}}
    & Ours & 
    \textbf{32.37 $\pm$ 0.22} & \textbf{32.78 $\pm$ 0.22} & \textbf{33.03 $\pm$ 0.20} &\textbf{32.97 $\pm$ 0.15} & \textbf{32.61 $\pm$ 0.13}\\
    & MLE. init. & 
    30.93 $\pm$ 0.28 & 31.43 $\pm$ 0.29 & 31.69 $\pm$ 0.28 & 32.04 $\pm$ 0.24 & 31.92 $\pm$ 0.20\\
    & Rand. init. & 
    31.61 $\pm$ 0.26 & 31.97 $\pm$ 0.27 & 32.21 $\pm$ 0.25 & 32.15 $\pm$ 0.23 & 31.92 $\pm$ 0.20\\
    & Zero. init. & 
    31.53 $\pm$ 0.26 & 32.02 $\pm$ 0.27 & 32.16 $\pm$ 0.26 & 32.03 $\pm$ 0.23 & 32.01 $\pm$ 0.20\\
    \cmidrule{2-7}
    \multirow{4}{*}{\rotatebox[origin=c]{90}{OOD}}
    & Ours & 
    \textbf{26.77 $\pm$ 2.28} & \textbf{27.18 $\pm$ 2.34} & \textbf{27.80 $\pm$ 2.20} & \textbf{28.67 $\pm$ 1.95} & \textbf{28.29 $\pm$ 1.71}\\
    & MLE. init. & 
    24.94 $\pm$ 2.40 & 25.05 $\pm$ 2.49 & 26.48 $\pm$ 2.29 & 27.93 $\pm$ 2.34 & 27.63 $\pm$ 2.18\\
    & Rand. init. & 
    26.25 $\pm$ 2.39 & 26.46 $\pm$ 2.38 & 27.70 $\pm$ 2.49 & 28.03 $\pm$ 2.33 & 27.62 $\pm$ 2.12\\
    & Zero. init. & 
    23.85 $\pm$ 2.11 & 26.09 $\pm$ 2.36 & 27.81 $\pm$ 2.47 & 28.09 $\pm$ 2.29 & 27.57 $\pm$ 2.13\\
    \bottomrule
    \end{tabular}
    }
    \end{center}
\end{table*}

\begin{figure*}[!tb]
    \begin{center}

    \resizebox{\linewidth}{!}{
    \renewcommand{\tabcolsep}{2pt}
    \def\figwidth{0.07\linewidth}%
    \newcommand{\authornote}[1]{
        \footnotesize 
        \adjustbox{rotate=90}{\parbox{\figwidth}{\centering #1}}
    }
    \begin{tabular}{*{11}{c}}
      \authornote{Ground Truth}     &  
        \loadimage{figures/appendix/denoise/realnvp}{width=\figwidth}{0}{10}\\
      \authornote{Observed Images}     & 
        \loadimage{figures/appendix/denoise/realnvp}{width=\figwidth}{10}{10}\\
      \authornote{Ours}        & 
        \loadimage{figures/appendix/denoise/realnvp}{width=\figwidth}{20}{10}\\
      \authornote{MLE init.}        & 
        \loadimage{figures/appendix/denoise/realnvp}{width=\figwidth}{30}{10}\\
      \authornote{Rand. init.}          &  
        \loadimage{figures/appendix/denoise/realnvp}{width=\figwidth}{40}{10}\\
      \authornote{Zero. init.}         &  
        \loadimage{figures/appendix/denoise/realnvp}{width=\figwidth}{50}{10}\\
      & \multicolumn{5}{c}{\fbox{\parbox{\calctotalwidth{\figwidth}{5}}{\centering Test}}} 
      & \multicolumn{5}{c}{\fbox{\parbox{\calctotalwidth{\figwidth}{5}}{\centering OOD}}}
    \end{tabular}
    }
    \caption{
    Results of solving denoising tasks on CelebA faces and out-of-distribution images with RealNVP as the generative prior. Hyperparameter $\lambda = 0.5$.
    }
    \label{fig:denoise-realnvp-full}
    \end{center}
\end{figure*}

\begin{figure*}[!tb]
    \begin{center}

    \resizebox{\linewidth}{!}{
    \renewcommand{\tabcolsep}{2pt}
    \def\figwidth{0.07\linewidth}%
    \newcommand{\authornote}[1]{
        \footnotesize 
        \adjustbox{rotate=90}{\parbox{\figwidth}{\centering #1}}
    }
    \begin{tabular}{*{11}{c}}
      \authornote{Ground Truth}     &  
        \loadimage{figures/appendix/denoise/glow}{width=\figwidth}{0}{10}\\
      \authornote{Observed Images}     & 
        \loadimage{figures/appendix/denoise/glow}{width=\figwidth}{10}{10}\\
      \authornote{Ours}        & 
        \loadimage{figures/appendix/denoise/glow}{width=\figwidth}{20}{10}\\
      \authornote{MLE init.}        & 
        \loadimage{figures/appendix/denoise/glow}{width=\figwidth}{30}{10}\\
      \authornote{Rand. init.}          &  
        \loadimage{figures/appendix/denoise/glow}{width=\figwidth}{40}{10}\\
      \authornote{Zero. init.}         &  
        \loadimage{figures/appendix/denoise/glow}{width=\figwidth}{50}{10}\\
      & \multicolumn{5}{c}{\fbox{\parbox{\calctotalwidth{\figwidth}{5}}{\centering Test}}} 
      & \multicolumn{5}{c}{\fbox{\parbox{\calctotalwidth{\figwidth}{5}}{\centering OOD}}}
    \end{tabular}
    }
    \caption{
    Results of solving denoising tasks on CelebA faces and out-of-distribution images with GLOW as the generative prior. Hyperparameter $\lambda = 0.3$.
    }
    \label{fig:denoise-glow-full}
    \end{center}
\end{figure*}


\begin{figure*}[!tb]
    \begin{center}

    \resizebox{\linewidth}{!}{
    \renewcommand{\tabcolsep}{2pt}
    \def\figwidth{0.07\linewidth}%
    \newcommand{\authornote}[1]{
        \footnotesize 
        \adjustbox{rotate=90}{\parbox{\figwidth}{\centering #1}}
    }
    \begin{tabular}{*{11}{c}}
      \authornote{Ground Truth}     &  
        \loadimage{figures/appendix/csense4000/realnvp}{width=\figwidth}{0}{10}\\
      \authornote{Ours}        & 
        \loadimage{figures/appendix/csense4000/realnvp}{width=\figwidth}{20}{10}\\
      \authornote{MLE init.}        & 
        \loadimage{figures/appendix/csense4000/realnvp}{width=\figwidth}{30}{10}\\
      \authornote{Rand. init.}          &  
        \loadimage{figures/appendix/csense4000/realnvp}{width=\figwidth}{40}{10}\\
      \authornote{Zero. init.}         &  
        \loadimage{figures/appendix/csense4000/realnvp}{width=\figwidth}{50}{10}\\
      & \multicolumn{5}{c}{\fbox{\parbox{\calctotalwidth{\figwidth}{5}}{\centering Test}}} 
      & \multicolumn{5}{c}{\fbox{\parbox{\calctotalwidth{\figwidth}{5}}{\centering OOD}}}
    \end{tabular}
    }
    \caption{
        Results of solving NCS when $m=4000$ on CelebA faces and out-of-distribution images with RealNVP as the generative prior. Hyperparameter $\lambda = 0.5$.
    }
    \label{fig:ncs4000-realnvp-full}
    \end{center}
\end{figure*}

\begin{figure*}[!tb]
    \begin{center}

    \resizebox{\linewidth}{!}{
    \renewcommand{\tabcolsep}{2pt}
    \def\figwidth{0.07\linewidth}%
    \newcommand{\authornote}[1]{
        \footnotesize 
        \adjustbox{rotate=90}{\parbox{\figwidth}{\centering #1}}
    }
    \begin{tabular}{*{11}{c}}
      \authornote{Ground Truth}     &  
        \loadimage{figures/appendix/csense4000/glow}{width=\figwidth}{0}{10}\\
      \authornote{Ours}        & 
        \loadimage{figures/appendix/csense4000/glow}{width=\figwidth}{20}{10}\\
      \authornote{MLE init.}        & 
        \loadimage{figures/appendix/csense4000/glow}{width=\figwidth}{30}{10}\\
      \authornote{Rand. init.}          &  
        \loadimage{figures/appendix/csense4000/glow}{width=\figwidth}{40}{10}\\
      \authornote{Zero. init.}         &  
        \loadimage{figures/appendix/csense4000/glow}{width=\figwidth}{50}{10}\\
      & \multicolumn{5}{c}{\fbox{\parbox{\calctotalwidth{\figwidth}{5}}{\centering Test}}} 
      & \multicolumn{5}{c}{\fbox{\parbox{\calctotalwidth{\figwidth}{5}}{\centering OOD}}}
    \end{tabular}
    }
    \caption{
    Results of solving NCS when $m=4000$ on CelebA faces and out-of-distribution images with GLOW as the generative prior. Hyperparameter $\lambda = 0.5$.
    }
    \label{fig:ncs4000-glow-full}
    \end{center}
\end{figure*}


\begin{figure*}[!tb]
    \begin{center}

    \resizebox{\linewidth}{!}{
    \renewcommand{\tabcolsep}{2pt}
    \def\figwidth{0.07\linewidth}%
    \newcommand{\authornote}[1]{
        \footnotesize 
        \adjustbox{rotate=90}{\parbox{\figwidth}{\centering #1}}
    }
    \begin{tabular}{*{11}{c}}
      \authornote{Ground Truth}     &  
        \loadimage{figures/appendix/csense2000/realnvp}{width=\figwidth}{0}{10}\\
      \authornote{Ours}        & 
        \loadimage{figures/appendix/csense2000/realnvp}{width=\figwidth}{20}{10}\\
      \authornote{MLE init.}        & 
        \loadimage{figures/appendix/csense2000/realnvp}{width=\figwidth}{30}{10}\\
      \authornote{Rand. init.}          &  
        \loadimage{figures/appendix/csense2000/realnvp}{width=\figwidth}{40}{10}\\
      \authornote{Zero. init.}         &  
        \loadimage{figures/appendix/csense2000/realnvp}{width=\figwidth}{50}{10}\\
      & \multicolumn{5}{c}{\fbox{\parbox{\calctotalwidth{\figwidth}{5}}{\centering Test}}} 
      & \multicolumn{5}{c}{\fbox{\parbox{\calctotalwidth{\figwidth}{5}}{\centering OOD}}}
    \end{tabular}
    }
    \caption{
        Results of solving NCS when $m=2000$ on CelebA faces and out-of-distribution images with RealNVP as the generative prior. Hyperparameter $\lambda = 0.3$.
    }
    \label{fig:ncs2000-realnvp-full}
    \end{center}
\end{figure*}

\begin{figure*}[!tb]
    \begin{center}

    \resizebox{\linewidth}{!}{
    \renewcommand{\tabcolsep}{2pt}
    \def\figwidth{0.07\linewidth}%
    \newcommand{\authornote}[1]{
        \footnotesize 
        \adjustbox{rotate=90}{\parbox{\figwidth}{\centering #1}}
    }
    \begin{tabular}{*{11}{c}}
      \authornote{Ground Truth}     &  
        \loadimage{figures/appendix/csense2000/glow}{width=\figwidth}{0}{10}\\
      \authornote{Ours}        & 
        \loadimage{figures/appendix/csense2000/glow}{width=\figwidth}{20}{10}\\
      \authornote{MLE init.}        & 
        \loadimage{figures/appendix/csense2000/glow}{width=\figwidth}{30}{10}\\
      \authornote{Rand. init.}          &  
        \loadimage{figures/appendix/csense2000/glow}{width=\figwidth}{40}{10}\\
      \authornote{Zero. init.}         &  
        \loadimage{figures/appendix/csense2000/glow}{width=\figwidth}{50}{10}\\
      & \multicolumn{5}{c}{\fbox{\parbox{\calctotalwidth{\figwidth}{5}}{\centering Test}}} 
      & \multicolumn{5}{c}{\fbox{\parbox{\calctotalwidth{\figwidth}{5}}{\centering OOD}}}
    \end{tabular}
    }
    \caption{
    Results of solving NCS when $m=2000$ on CelebA faces and out-of-distribution images with GLOW as the generative prior. Hyperparameter $\lambda = 0.3$.
    }
    \label{fig:ncs2000-glow-full}
    \end{center}
\end{figure*}


\begin{figure*}[!tb]
    \begin{center}

    \resizebox{\linewidth}{!}{
    \renewcommand{\tabcolsep}{2pt}
    \def\figwidth{0.07\linewidth}%
    \newcommand{\authornote}[1]{
        \footnotesize 
        \adjustbox{rotate=90}{\parbox{\figwidth}{\centering #1}}
    }
    \begin{tabular}{*{11}{c}}
      \authornote{Ground Truth}     &  
        \loadimage{figures/appendix/inpaint/realnvp}{width=\figwidth}{0}{10}\\
      \authornote{Observed Images}     & 
        \loadimage{figures/appendix/inpaint/realnvp}{width=\figwidth}{10}{10}\\
      \authornote{Ours}        & 
        \loadimage{figures/appendix/inpaint/realnvp}{width=\figwidth}{20}{10}\\
      \authornote{MLE init.}        & 
        \loadimage{figures/appendix/inpaint/realnvp}{width=\figwidth}{30}{10}\\
      \authornote{Rand. init.}          &  
        \loadimage{figures/appendix/inpaint/realnvp}{width=\figwidth}{40}{10}\\
      \authornote{Zero. init.}         &  
        \loadimage{figures/appendix/inpaint/realnvp}{width=\figwidth}{50}{10}\\
      & \multicolumn{5}{c}{\fbox{\parbox{\calctotalwidth{\figwidth}{5}}{\centering Test}}} 
      & \multicolumn{5}{c}{\fbox{\parbox{\calctotalwidth{\figwidth}{5}}{\centering OOD}}}
    \end{tabular}
    }
    \caption{
        Results of inpainting CelebA faces and out-of-distribution images with RealNVP as the generative prior. 
        Hyperparameter $\lambda=1.5$.
    }
    \label{fig:inpaint-realnvp-full}
    \end{center}
\end{figure*}

\begin{figure*}[!tb]
    \begin{center}

    \resizebox{\linewidth}{!}{
    \renewcommand{\tabcolsep}{2pt}
    \def\figwidth{0.07\linewidth}%
    \newcommand{\authornote}[1]{
        \footnotesize 
        \adjustbox{rotate=90}{\parbox{\figwidth}{\centering #1}}
    }
    \begin{tabular}{*{11}{c}}
      \authornote{Ground Truth}     &  
        \loadimage{figures/appendix/inpaint/glow}{width=\figwidth}{0}{10}\\
      \authornote{Observed Images}     & 
        \loadimage{figures/appendix/inpaint/glow}{width=\figwidth}{10}{10}\\
      \authornote{Ours}        & 
        \loadimage{figures/appendix/inpaint/glow}{width=\figwidth}{20}{10}\\
      \authornote{MLE init.}        & 
        \loadimage{figures/appendix/inpaint/glow}{width=\figwidth}{30}{10}\\
      \authornote{Rand. init.}          &  
        \loadimage{figures/appendix/inpaint/glow}{width=\figwidth}{40}{10}\\
      \authornote{Zero. init.}         &  
        \loadimage{figures/appendix/inpaint/glow}{width=\figwidth}{50}{10}\\
      & \multicolumn{5}{c}{\fbox{\parbox{\calctotalwidth{\figwidth}{5}}{\centering Test}}} 
      & \multicolumn{5}{c}{\fbox{\parbox{\calctotalwidth{\figwidth}{5}}{\centering OOD}}}
    \end{tabular}
    }
    \caption{
    Results of inpainting CelebA faces and out-of-distribution images with GLOW as the generative prior. 
    Hyperparameter $\lambda=1.5$.
    }
    \label{fig:inpaint-glow-full}
    \end{center}
\end{figure*}